# Improving Global Forest Mapping by Semi-automatic Sample Labeling with Deep Learning on Google Earth Images

Qian Shi[1], Xiaolei Qin[1], Lingyu Sun[1], Zitao Shen[1], Xiaoping Liu[1]*, Xiaocong Xu[1], Jiaxin Tian[1], Rong Liu[1], Andrea Marinoni[2, 3]

*1 Guangdong Provincial Key Laboratory for Urbanization and Geo-simulation, School of Geography and Planning, Sun Yat-sen University*

*2 Dept. of Physics and Technology, UiT the Arctic University of Norway, Tromsø, Norway*

*3 Department of Engineering, University of Cambridge, Cambridge, UK*

*\*corresponding author: liuxp3@mail.sysu.edu.cn*

**Abstract**

    Global forest cover is critical to the provision of certain ecosystem services. With the advent of the google earth engine cloud platform, fine resolution global land cover mapping task could be accomplished in a matter of days instead of years. The amount of global forest cover (GFC) products has been steadily increasing in the last decades. However, it's hard for users to select suitable one due to great differences between these products, and the accuracy of these GFC products has not been verified on global scale. To provide guidelines for users and producers, it is urgent to produce a validation sample set at the global level. However, this labeling task is time and labor-consuming, which has been the main obstacle to the progress of global land cover mapping. In this research, a labor-efficient semi-automatic framework is introduced to build a biggest ever Forest Sample Set (FSS) contained 395280 scattered samples

categorized as forest, shrubland, grassland, impervious surface, etc. On the other hand, to provide guidelines for the users, we comprehensively validated the local and global mapping accuracy of all existing 30m GFC products, and analyzed and mapped the agreement of them. Moreover, to provide guidelines for the producers, optimal sampling strategy was proposed to improve the global forest classification. Furthermore, a new global forest cover named GlobeForest2020 has been generated, which proved to improve the previous highest state-of-the-art accuracies (obtained by Gong et al., 2017) by 2.77% in uncertain grids and by 1.11 % in certain grids.

## I. Introduction

Forest is an important part of ecosystem, as it is significant to biodiversity, climate regulation, carbon storage and water supply (Shi, Zhao, Tang, & Fang, 2011; Zeng et al., 2018; Xiaomei Zhang et al., 2020; Zhang et al., 2017). Monitoring global forest cover change using remote sensing technology is essential to studies on climate change, carbon reserve change and human activities (Cadman, 2011; Hansen et al., 2013; Liu et al., 2021; Xiaomei Zhang et al., 2020). Therefore, global forest cover dominates in all global land cover (GLC) products (Hansen et al., 2013; Xiaomei Zhang et al., 2020).

In the early stage, land-cover products were mainly generated manually by visual interpretation, which is time consuming and labor intensive, especially when performed at global scale. For example, Chinese Academy of Sciences (CAS) and The Ministry of Agriculture took three years to investigate the National Resources in China (Jiyuan, 1997). Improvements of image processing algorithm, computing power and data volume propelled researches on machine learning based automatic classification for land cover mapping (Hagner & Reese, 2007; Hansen et al., 2013; Jia, Li, Shi, & Zhu, 2019; Pelletier, Valero, Inglada, Champion, & Dedieu, 2016; P. Potapov, Turubanova, & Hansen, 2011; P. V. Potapov et al., 2012; Spracklen & Spracklen, 2021). To improve accuracy of land cover mapping, methods including multi-source data fusion, ensemble learning, object-based classification, post-processing, and transfer learning were exploited by previous studies (Dorren, Maier, & Seijmonsbergen, 2003; Kasetkasem, Arora, & Varshney, 2005; Knauer et al., 2019; X. Li, Du, & Ling, 2013; Qin et al., 2015; Solberg, Taxt, & Jain, 1996; Van Coillie, Verbeke, & De Wulf, 2007; Zhang et al., 2017).

Recently, GLC products for forest cover mapping have been generated by automatic remote sensing image classification. Early GLC products are mainly based on Moderate solution Imaging Spectroradiometer (MODIS) data, ENVISAT-MERIS data and other coarse-resolution data, e.g., Global Land-cover Classification (GLC_2000) product at 1 km (Hansen, DeFries, Townshend, & Sohlberg, 2000), Land-cover product (MOD12Q1) at 500m (Friedl et al., 2010), the European Space Agency (ESA) Climate Change Initiative Land-cover product (CCI_LC) (Lamarche et al., 2017) and Global Land-cover map (GlobCover) at 300m (Defourny et al., 2006). Due to the free available Landsat Thematic Mapper (TM) and Enhanced Thematic Mapper+ (ETM+) data from 2011, 30m resolution GLC products quickly predominated in GLC monitoring task (Chen et al., 2015; Gong et al., 2013; Hansen et al., 2013; Xiao Zhang, Liangyun Liu, Xidong Chen, et al., 2020; Xiao Zhang, Liangyun Liu,

Changshan Wu, et al., 2020; Xiaomei Zhang et al., 2020). Efforts were made to improve accuracy of GLC products over the last few years. FROM_GLC (Gong et al., 2013) was produced by employing four classifiers using Landsat TM and ETM+ data, and MODIS enhanced vegetation index (MODIS EVI) time series data was also referenced in developing the training and test sample databases. GlobeLand30 (Chen et al., 2015) was produced by classifying Landsat-like satellite images using an approach based on the integration of pixel- and object-based methods with knowledge (POK-based). GLC_FCS30 (Xiao Zhang, Liangyun Liu, Xidong Chen, et al., 2020; Xiao Zhang, Liangyun Liu, Changshan Wu, et al., 2020) was produced by building a local adaptive random forest model using Global Spatial Temporal Spectra Library (GSPECLib) and time series of Landsat imagery.

Some studies also focused solely on forest, in order to obtain more accurate forest cover product. GLADForest (Hansen et al., 2013) is a 30m resolution global forest cover map product including percent tree cover in 2000, annual forest loss from 2000 to 2019, and forest gain from 2000 to 2012, which was produced by building a regression model using growing season composite Landsat-7 ETM+ image data to estimate the maximum tree canopy cover. GFC30 (Xiaomei Zhang et al., 2020) is a 30m resolution global forest cover map product, which was produced by building random forest classifier to classify each forest ecological zone using time series of Landsat 8 image data. It is worth noting that the generation of many GLC products mentioned above benefits from the advent of remote sensing cloud platforms, e.g., Google Earth engine (GEE) which integrates various remote sensing datasets and advanced image processing and classification algorithms (Xiao Zhang, Liangyun Liu, Xidong Chen, et al., 2020; Xiao Zhang, Liangyun Liu, Changshan Wu, et al., 2020; Xiaomei Zhang et al., 2020).

As a result, producing time of fine-resolution GLC products could be dramatically shortened.

Nevertheless, growing number of forest cover products does not mean more accurate surface forest distribution monitoring. These GFC products represent great difference. Agreement of GLC products varies by region, while significantly uncertain regions also play a critical role in ecosystem (Liu et al., 2021). One possible reason of disagreement is the definition difference of forest among products. Tree-cover percentage (TCP) and tree height (TH) are most common criteria (Hansen et al., 2013). For the threshold of TCP, 15% is considered by GLC_FCS30 (Xiao Zhang, Liangyun Liu, Xidong Chen, et al., 2020; Xiao Zhang, Liangyun Liu, Changshan Wu, et al., 2020) and FROM_GLC (Gong et al., 2013), while 10% is used for GlobeLand30 and GFC30. At the same time, FROM_GLC has a TH threshold of 3 m, while 5 m is used by GLADForest and GFC30. Lower threshold of TCP and TH means more samples would to be classified as forest. However, forest area statistics on the above-mentioned products violates that expectation. Forest area of GlobeLand30 and GFC30 (TCP:10%) is 30,537,642 km² and 28,529,533 km², both lower than either FROM_GLC or GLC_FCS30 (TCP:15%) with 36,547,005 km² and 43,305,677 km² respectively. Therefore, definition of forest is an unreliable reference for users to select GLC product.

To establish an equitable reference for GLC product users and producers, a third-party accuracy evaluation for GLC products is necessary. This implies to build a consistent and rigorous validation sample set. However, annotating forest sample set is a tedious work, requiring expertise and large amount of manpower, which falls far short of the speed of product generation. Thus, the bottleneck of this procedure is represented by the identification of a design strategy that enables to generate high-quality validation sample set with minimal labor and time costs. Furthermore, improving the accuracy of the forest mapping product is also a hot topic. By validating existing products, we found significant disagreement in transition areas between humid and arid zones, where the forest showed up sparsely and mixed with other land covers such as bare land, shrubland and grassland. In those areas, the mixed pixel problem (i.e., spectrum of one single pixel is the mixture of several different landcover spectra) brings challenges for accurate forest cover mapping. Moreover, tree species in those transition areas are mainly coniferous forests and mingled forests, whose spectra severely confused with shrubs and grasslands. Thus, how to improve accuracy in controversial regions is the breakthrough point to produce more accurate global forest cover product.

To solve the aforesaid scientific problems, we first built a semi-automatic sample annotating mechanism that generates high quality samples especially from controversial regions in a time and labor efficient way. Although these vegetation classes (forest, shrub, grassland, etc.) show a slight difference in the spectral signatures on the Landsat imagery, a distinct difference on the texture distribution on Very High Resolution (VHR) Imagery can be recorded. Thus, we fully utilized deep learning (DL) classification model to exploit the texture structure of different vegetation types. The DL model actively provides initial labels for geocoded VHR patches to assist human annotator, and get re-trained with new labels after human examination. While DL model can deliver labels accurately, samples having high confidence need less human labeling costs. Leveraging this mechanism, 51.33%-77% manual work can be saved. We produced 395280 scattered samples in total, distributed worldwide by stratified sampling, classified as forest, shrubland, grassland, cropland, impervious surface, and other classes. This dataset with geographic coordinates has been published online to provide validation for researchers around the world. Furthermore, to get more satisfying classification map, diverse sample strategies were discussed to evaluate the effectiveness to produce high precision GFC products.

This paper is divided into six parts: section II reviews the current global 30m GFC products, section III describes the sample annotating mechanism and classification strategies with produced samples, section IV presents the results of accuracy enhancement of GlobeForest2020, section V conducts discussions and section VI summarizes this paper. Our contribution includes: (1) a semi-automatic sample labeling mechanism that actively labels most inconsistent sample set with minimum manpower costs and highest labeling quality has been implemented; (2) A biggest ever forest sample set (FFS) for the training and validation of GFC products have been provided in the global scale; (3) a new global forest cover product named GlobeForest2020 has been generated, which proved to improve the previous highest state-of-the-art accuracies (obtained by Gong et al., 2017) by 2.77% in uncertain grids and by 1.11 % in certain grids.

## II. Review and Comparative Analysis of the Current 30m GFC Products

In this part, the disagreement of different products is systematically analyzed. In this study, five global 30m GFC products were reviewed and investigated, including the forest layer of three GLC products (FROM_GLC30 2017, GlobeLand30 2020, and GLC_FCS30-2020) ([Gong et al., 2019](#); [Jun, Ban, & Li, 2014](#); [Xiao Zhang, Liangyun Liu, Xidong Chen, et al., 2020](#)), and two GFC products (GFC30 2018 and GLADForest 2019) ([Hansen et al., 2013](#);

Xiaomei Zhang et al., 2020).

In order to intuitively present the spatial agreement between the different products, we utilize spatial superposition method to generate the pixel-by-pixel spatial correspondence for five GFC products. The agreement value is used to record the times of all GFC products voted to the forest class. The agreement of a pixel takes integer values in the range [0,5]. The highest value equals to 5, which means all products voted to the forest class. On the contrary, the lowest value equals to 0, which means all products voted to the non-forest class. If the agreement value equals to 2 or 3, it means the pixel represents low agreement on whether belongs to the forest class or not. The spatial agreement distribution for all five products and for each pair of products is given in Figure 1. From the forest products agreement map shown in Figure 1, we can find that a high value of inconsistency is identified by a high density of red dots in a given area.

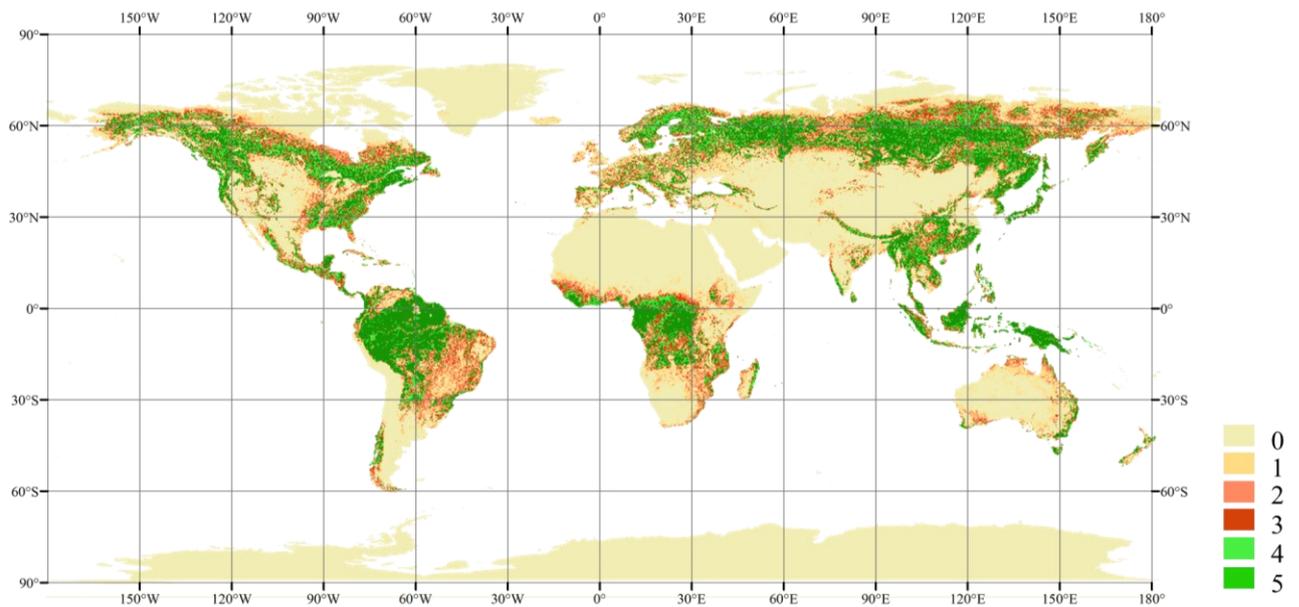

Figure 1 Forest products agreement map (If a pixel equals to 2 or 3, it shows low certainty to be the "forest" class. If a pixel value equals to 0 or 1, it shows high certainty to be the "non-forest" class. If a pixel value equals to 4 or 5, it shows high certainty to be the "forest" class).

The uncertainty of any region is measured as the ratio of the area of pixels with values of 2 or 3 to the area of pixels with values not less than 1 within the region, which can be represented by equation (1):

$$Uncertainty\_23^j = \frac{COUNT\left(C^j_{i\ value\in[2,3]}\right)}{COUNT\left(C^j_{k\ value\in[1,2,3,4,5]}\right)} \quad (1)$$

where $C^j_{i\ value\in[2,3]}$ is the $i^{th}$ pixel in the $j^{th}$ region with value of 2 or 3, $C^k_{i\ value\in[1,2,3,4,5]}$ is the $k^{th}$ pixel in the $j^{th}$ region with value of 1, 2, 3, 4 or 5.

A threshold for $Uncertainty\_23^j$ is used to differentiate the **certain regions and uncertain regions.** The accuracy evaluation and forest classification are conducted in the unit of 5° grid. And the certainty is conducted in grids that the area of pixels with value over 1 account for more than 10% of the total area. In this study, the threshold of 0.3 is chosen, leading to a selection of 160 grids considered as uncertain grids (Figure 2). They are mainly distributed in northern and southern North America, eastern and southern South America, central Africa, the Mediterranean region, northern and southern Asia, and northern and southern Australia.

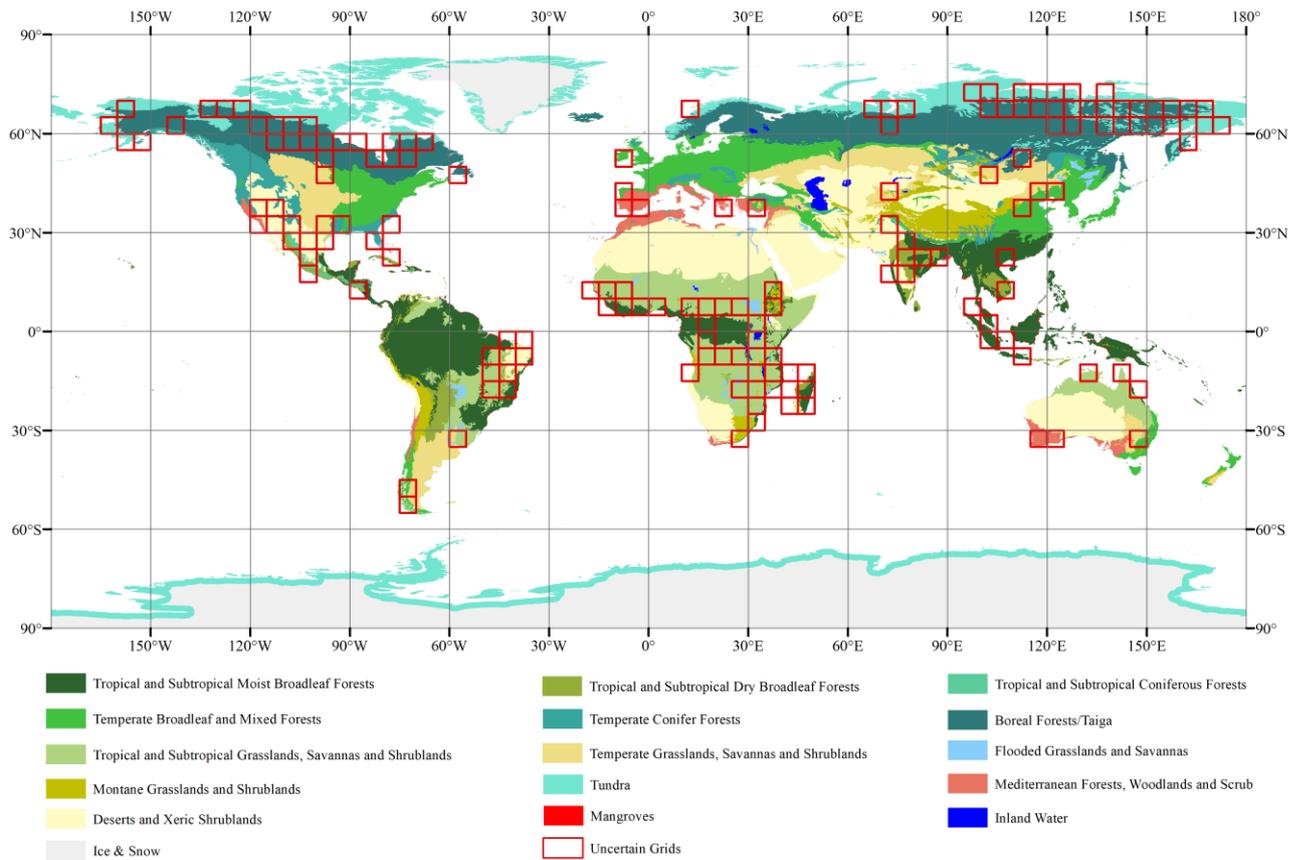

Figure 2 Uncertain grids distribution.

*Uncertainty_23* of continents and ecoregions are shown in Table 1. Africa has the highest *Uncertainty_23* of 0.3562, and followed by North America with 0.3371. They have the highest proportion of grids divided into uncertain ones (Figure 2). As for ecoregions (Table 2), Tropical and Subtropical Coniferous Forests have the highest *Uncertainty_23* of 0.4483 and followed by Tropical and Subtropical Grasslands, Savannas and Shrublands (0.4349), Mediterranean Forests, Woodlands and Scrub (0.4059), Tropical and Subtropical Dry Broadleaf Forests (0.4046), Flooded Grasslands and Savannas (0.4019). Ecoregion of Temperate Grasslands, Savannas and Shrublands has low *Uncertainty_23*.

Table 1 Uncertainty of continents.

| Continent | *Uncertainty_23* |
|---|---|
| Africa | 0.3562 |
| Australia | 0.2761 |
| Asia | 0.2902 |
| Europe | 0.3209 |
| North America | 0.3371 |
| South America | 0.3003 |

Table 2 Uncertainty of ecoregions.

| Ecoregion | *Uncertainty_23* |
|---|---|
| Tropical and Subtropical Moist Broadleaf Forests | 0.2052 |
| Tropical and Subtropical Dry Broadleaf Forests | 0.4046 |
| Tropical and Subtropical Coniferous Forests | 0.4483 |
| Temperate Broadleaf and Mixed Forests | 0.3230 |
| Temperate Conifer Forests | 0.3132 |
| Boreal Forests/Taiga | 0.3029 |
| Tropical and Subtropical Grasslands, Savannas and Shrublands | 0.4349 |
| Temperate Grasslands, Savannas and Shrublands | 0.2954 |
| Flooded Grasslands and Savannas | 0.4019 |
| Montane Grasslands and Shrublands | 0.3896 |
| Tundra | 0.3274 |
| Mediterranean Forests, Woodlands and Scrub | 0.4059 |
| Deserts and Xeric Shrublands | 0.3173 |
| Mangroves | 0.3855 |
| Inland Water | 0.2355 |
| Ice & Snow | 0.2250 |

Scatter plots of the 5 GCF products after aggregation to a spatial resolution of 0.5° were used to analyze the spatial agreement between the arbitrary two different global forest products, and results are displayed in Figure 3. The products pairs have lower agreement in the uncertain grids. In the uncertain grids, the best agreement appears between GFC30 2018 and GLADForest 2019 with coefficient of 0.951, $R^2$ of 0.715 and *RMSE* of 0.146. This is followed by FROM_GLC30 2017 against GFC30 2018 (*coefficient* = 0.821, $R^2$ = 0.693 *RMSE* = 0.133) and GLC_FCS30-2020 against GFC30 2018 (*coefficient* = 0.797, $R^2$ = 0.421, *RMSE* = 0.228). The distribution of points is more dispersed in the plot of GLADForest 2019 against GlobeLand30 2020, GLC_FCS30-2020 against FROM_GLC30 2017, GLADForest 2019 against GLC_FCS30-2020, which illustrates these forest map pairs have less spatial agreement. As for certain grids, the best agreement appears between GFC30 2018 and FROM_GLC30 2017 with coefficient of 0.991, $R^2$ of 0.896 and *RMSE* of 0.108, followed by GFC30 2018 against GLADForest 2019 with coefficient of 0.972, $R^2$ of 0.905 and *RMSE* of 0.100. FROM_GLC30 2017 has high agreement against GLADForest 2019 (*coefficient* = 0.926, $R^2$ =0.899 and *RMSE* = 0.103) and against GlobeLand30 2020 (*coefficient* = 0.913, $R^2$ =0.790 and *RMSE* = 0.152).

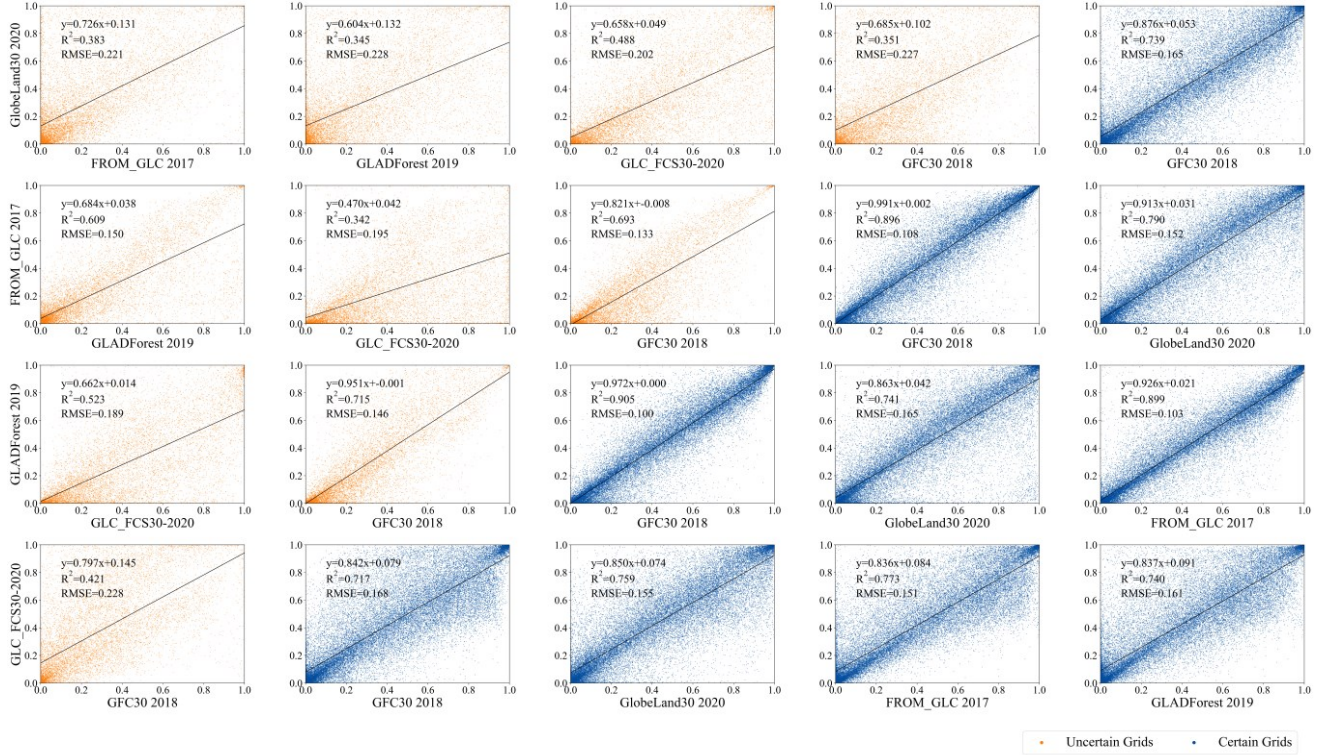

Figure 3 Scatter plots between the any two of the 5 global forest products obtained using a 0. 5° × 0.5° grid. The axes represent the fraction of forest within each 0. 5° × 0.5° grid cell. The orange points are from uncertain grids and the blue points are from certain grids.

Then, we computed the proportion of cropland and shrubland samples confused with forest, and results are reported in Table 3. It is measured by the index *ratio_mis$_{class}$*, which is defined as below:

$$ratio\_mis_{class} = \frac{Count(P_{class}^{forest})}{Count(P_{class})} \quad (2)$$

where $P_{class}$ refers to the true type of class, and $P_{class}^{forest}$ refers to the point of certain class classified to forest.

The highest proportion of confused categories in each ecoregion is shrubs, followed by cropland. In Temperate Broadleaf and Mixed Forests and Temperate Conifer Forests, the proportion of confused with shrubs is relatively small, while the proportion of misclassification of cropland is relatively large. Because the spectral characteristics of shrubland is the closest to forest, the likelihood of misclassification is very high.

Table 3 Proportion of confused classes.

| Ecoregion | $ratio\_mis_{Cropland}$ | $ratio\_mis_{Shrubland}$ |
|---|---|---|
| Tropical and Subtropical Moist Broadleaf Forests | 0.1106 | 0.8894 |
| Tropical and Subtropical Dry Broadleaf Forests | 0.0736 | 0.9264 |
| Tropical and Subtropical Coniferous Forests | 0.0029 | 0.9971 |
| Temperate Broadleaf and Mixed Forests | 0.3557 | 0.6443 |
| Temperate Conifer Forests | 0.3285 | 0.6715 |
| Boreal Forests/Taiga | 0.0202 | 0.9798 |
| Tropical and Subtropical Grasslands, Savannas and Shrublands | 0.0259 | 0.9741 |
| Temperate Grasslands, Savannas and Shrublands | 0.0676 | 0.9324 |
| Flooded Grasslands and Savannas | 0.0112 | 0.9888 |
| Montane Grasslands and Shrublands | 0.0782 | 0.9218 |
| Tundra | 0.0113 | 0.9887 |
| Mediterranean Forests, Woodlands and Scrub | 0.1838 | 0.8162 |
| Deserts and Xeric Shrublands | 0.0177 | 0.9823 |
| Mangroves | 0.1286 | 0.8714 |
| Inland Water | 0.0000 | 1.0000 |
| Ice & Snow | 0.0000 | 1.0000 |

## III. Methodology

This section describes our semi-automatic approach to produce labeled samples and how these samples are used for forest classification. In section 3.1, the definition of sample labeling rules, steps of semi-automatic labeling mechanism and distribution of FFS will be described in detail. In section 3.2, we will introduce the proposed sampling strategy to generated a more accurate classification map.

## 3.1 Semi-automatic Sample Labeling Mechanism

### 3.1.1 Annotating Rules and Obtained samples

High-resolution Google Earth Images (HR-GEI) are used to determine the type of sample points. Forest, shrubland, grassland, cropland, and impervious types are more distinguishable in HR-GEI compared with Landsat imagery due to higher resolution and more obvious spectral and texture features (W. Li et al., 2020). Figure 4 shows some examples of HR-GEI for different land cover types. Pixels whose tree-cover percentage is greater than 15% and tree height is greater than 3m will be defined as forest. Forest and shrubland can be distinguished by the height, shape and distribution of vegetation. In the aspect of height, the height of forest is generally high with obvious shadow, while the shrub is generally short with little shadow; in the aspect of shape, the forest vegetation is generally large, and the obvious crown structure can be seen in HR-GEI, while the shrub is generally small, and there is no obvious shape in HR-GEI; in the aspect of distribution, the forest is usually in patches, and the shrubs are scattered. Generally, forest is usually compact, and obvious height difference with the ground can be seen in HR-GEI. The local texture of forest is obvious, while shrub is generally sparse, and the height difference of surface and shrub is not obvious. Therefore, when labeling, if the vegetation has the characteristics of dense, obvious shadow, large shape and relatively high, it will be labeled as forest. If the vegetation has the characteristics of small shape, scattered, obvious surface between vegetation, low rate of large areas and obvious edge, it will be labeled as shrubland. Grassland is generally successive, flat in texture, low in height in HR-GEI. Farmland is relatively flat in height, and texture, and there are some ridges like lines in the farmland, which can be clearly distinguished according to the shape and boundary. Impervious surface generally contains buildings, cement, roads, and so on., which is easy to distinguish in HR-GEI. Images with too many clouds, too many shadows and abnormal color and some blurring images will not be marked. We have labeled 395280 scattered samples distributed worldwide, mainly including shrub, grassland, and cropland.

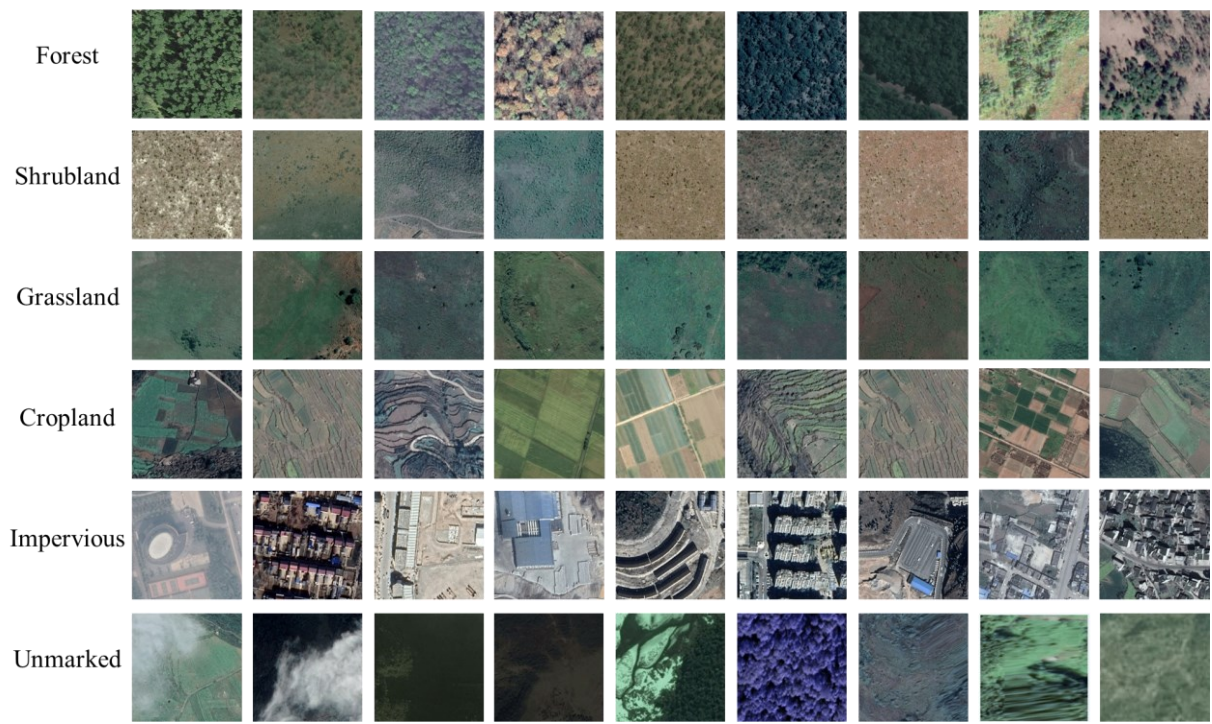

Figure 4 Examples of high-resolution Google Earth images of different land cover types.

## 3.1.2 Iterative Annotating Mechanism

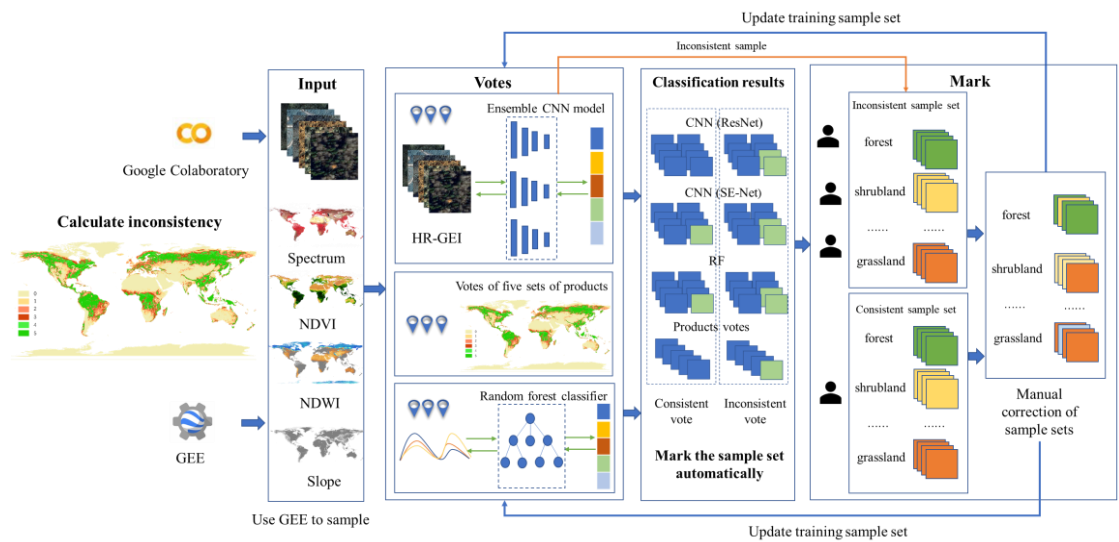

Figure 5 Flow of semi-automatic sample labeling mechanism.

At this point, we focus our attention on the problem of identification and labeling of samples showing high quality according to the discussion provided in the previous Section. To this aim, we built a labor efficient semi-automatic sample labeling mechanism to iteratively annotate high quality samples. The main way to reduce the workload is to classify the samples into hard and easy levels according to the interpretation difficulty. To guarantee the quality of the labeled sample set, the hard samples need more manual labeling costs, and vice versa. For a sample to be labeled, Landsat images and HR-GEI provide completely different description from spectral bands and spatial patches. We assume that a sample consistently predicted into one category by classifiers trained with different source images, belongs to that category with high confidence. This category can be marked as initial label, and the human just confirm the result, whose cost is smaller than labeling a new sample. Figure 5 shows the overview iterative process, which consists of three parts: sampling, manual annotating with different sample consistency and classifier updating. The whole process is summarized as follows. First, geo points are sampled with corresponding HR-GEI patches and feature vectors. The feature vector is derived from Landsat data and DEM data, including spectrum, NDVI ([Tucker, 1979](#)), NDWI ([Xu, 2006](#)) and slope. Second, using those sample points with initial labels from previous GLC products, multiple Convolutional Neural Networks (CNN) classifiers and random forest (RF) classifiers are trained, respectively. And next, the trained classifiers are used to predict on all sample points and generate votes with pseudo label. According to the votes obtained by the trained classifiers and five sets of products, sample points are divided into two sets (i.e., consistent set and inconsistent set) and re-annotated manually under different amount of labor. After re-annotating, CNN classifiers and random forest classifiers are re-trained with sample points and new labels. The entire process iterates several times until all sample points are annotated with correct labels.

The details of each part can be summarized as follows:

### A. Sampling

The sample points $\boldsymbol{p} = \{p\}$ are randomly generated with a NDVI-based stratified sampling method to ensure the diversity of sample type. For each sample point $p$, a feature vector $X^F$ is obtained from Google Earth Engine Landsat data collection and DEM data collection, and a HR-GEI patch $X^{GEI}$ is downloaded from Google Static Map API. $X^F$ contains spectrum, NDVI, NDWI and slope. On the other hand, $X^{GEI}$ is image patch in 1m ground resolution (i.e., 17$^{th}$ zoom level in Google Map) with 165×165 pixels, whose central geo coordinate equals to sample point coordinate. The initial label $y^{init}$ of each sample point is obtained from consistent votes of previous GLC and GFC products. In each iteration, we randomly select 10000 samples to be labeled.

## B. Manual annotating with different sample consistency

CNN classifiers $\{f_i^{CNN}(X^{GEI}, \theta_i^{CNN})\}_{i \in 1,2,..,N}$ are trained with $X^{GEI}$ and initial label $y^{init}$. At the same time, RF classifiers $\{f_i^{RF}(X^F, \theta_i^{RF})\}_{i \in 1,2,..,M}$ are trained with $X^F$ and initial label $y^{init}$. To avoid labeling error, the *K*-fold (*K*=8 is adopted in our experiments) strategy is adopted to train and validate multiple CNN classifiers and RF classifiers. ResNet (He, Zhang, Ren, & Jian, 2016) and SE-Net (Hu, Shen, & Sun, 2018) are used as the two kinds of neural networks to generate CNN classifiers, resulting 2*K* different CNN classifiers in total. And then, all sample points are predicted with classifiers $\{f_i^{CNN}(X^{GEI}, \theta_i^{CNN})\}_{i \in 1,2,..,2K}$ and $\{f_i^{RF}(X^F, \theta_i^{RF})\}_{i \in 1,2,..,M}$ (according to the notation that has been previously introduced) to generate new prediction set $\{\hat{y}_i\}_{i \in 1,2,...,2K+M}$. All votes of each sample point consist of the votes obtained by the trained classifiers and five sets of products, and the number of all votes of each sample point is $(2K + M + 5)$. Next, for each sample point, newly voted class $\hat{y}$ is assigned by the class with max votes $V^{class} = COUNT(Vote_{class})$, where $Vote_{class}$ is the vote whose type of voting is *class* and new votes $V$ are assigned with max votes $\max\{V^{class}\}_{class \in \{forest, non-forest.\}}$. According to new votes $V$, each sample $p_i$ is assigned to either the consistent set or the inconsistent set, expressed as below:

$$p_i \in \begin{cases} consistent & if\ V_i/(V_{num}) > \lambda \\ inconsistent & if\ V_i/(V_{num}) \leq \lambda \end{cases} \quad (3)$$

Where $V_{num}$ is the number of all votes ($V_{num} = 2K + M + 5$), $\lambda$ is manually assigned as a constant for the entire iterative annotating process ($\lambda = 0.9$ is adopted in our experiments).

Both consistent and inconsistent samples are uploaded to the online annotation system for manual annotating. Consistent samples $\boldsymbol{p_{con}} = \{p_i\}_{V_i/V_{num} > \lambda}$ will be marked only once.

Inconsistent samples $p_{incon} = \{p_i\}_{V_i/V_{num} \leq \lambda}$ will be marked three times, and voted for final decision. Since we have provided $\{\hat{y}\}$ voted by the trained classifiers and five sets of products, annotators only need to check if $\hat{y}$ is right and correct those wrongly predicted ones, which is more efficient than annotating from scratch. After manually annotating, we obtained $\{\hat{y}^{corrected}\}$.

C. **Classifier updating**

The corrected labels $\{\hat{y}^{corrected}\}$ will be used to update original labels $\{y^{init}\}$ of sample set $p = \{p\}$ and re-train the previous CNN classifiers $\{f_i^{CNN}(X^{GEI}, \theta_i^{CNN})\}_{i \in 1,2,...,2K}$ and random forest classifiers $\{f_i^{RF}(X^F, \theta_i^{RF})\}_{i \in 1,2,...,M}$, which makes their prediction results more accurate. As a result, fewer samples need to be corrected by human annotators in the new iteration.

### 3.1.3 Distribution of Labelled Samples

Table 4 and Table 5 show the number of samples in each ecoregion and continent. It is possible to appreciate that sample points are mainly concentrated in some ecoregions, such as Tropical and Subtropical Grasslands, Savannas and Shrublands and some Continents such as Africa, Europe, and Asia. The number of all forest points is 214720 and the number of all non-forest points is 180560. The number of all samples is 395280. The proportions of positive and negative samples are close to 1.

Table 4 Number of samples in each ecoregion.

| Ecoregions | forestPtsNum | nonforestPtsNum | allPtsNum |
|---|---|---|---|
| Tropical and Subtropical Moist Broadleaf Forests | 44109 | 19695 | 63804 |
| Tropical and Subtropical Dry Broadleaf Forests | 9071 | 6637 | 15708 |
| Tropical and Subtropical Coniferous Forests | 2209 | 851 | 3060 |
| Temperate Broadleaf and Mixed Forests | 37535 | 29436 | 66971 |
| Temperate Conifer Forests | 13816 | 7894 | 21710 |
| Boreal Forests/Taiga | 32164 | 17105 | 49269 |
| Tropical and Subtropical Grasslands, Savannas and Shrublands | 50117 | 60267 | 110384 |
| Temperate Grasslands, Savannas and Shrublands | 4431 | 9434 | 13865 |
| Flooded Grasslands and Savannas | 3333 | 3666 | 6999 |
| Montane Grasslands and Shrublands | 2840 | 4523 | 7363 |
| Tundra | 3345 | 5477 | 8822 |
| Mediterranean Forests, Woodlands, and Scrub | 3387 | 4626 | 8013 |
| Deserts and Xeric Shrublands | 6635 | 10151 | 16786 |
| Mangroves | 1705 | 472 | 2177 |
| Inland Water | 21 | 276 | 297 |
| Ice & Snow | 2 | 50 | 52 |
| sum | 214720 | 180560 | 395280 |

Note: forestPtsNum is the number of forest sample points, nonforestPtsNum is the number of non-forest sample points, allPtsNum is the number of all sample points.

Table 5 Number of samples in each continent.

| Continent | forestPtsNum | nonforestPtsNum | allPtsNum |
|---|---|---|---|
| Africa | 54307 | 60711 | 115018 |
| Europe | 38622 | 33029 | 71651 |
| Australia | 4058 | 4736 | 8794 |
| South America | 35454 | 25420 | 60874 |
| Asia | 49597 | 38257 | 87854 |
| North America | 32682 | 18407 | 51089 |
| sum | 214720 | 180560 | 395280 |

Note: forestPtsNum is the number of forest sample points, nonforestPtsNum is the number of non-forest sample points, allPtsNum is the number of all sample points.

This study builds a biggest ever Forest Sample Set (FSS) according to the iterative annotating mechanism. The samples in FSS are divided into training samples and validation samples according to the ratio of 3:7 (120098 training samples and 275182 validation samples). Figure 6 and Figure 7 show the distribution of training samples and validation samples. The training data are mainly concentrated in south-central North America, eastern South America, central Africa, mid- to high-latitude Europe and Asia, southern Asia and the margin of Australia. The distribution of the validation data is similar to the distribution pattern of the world's forests.

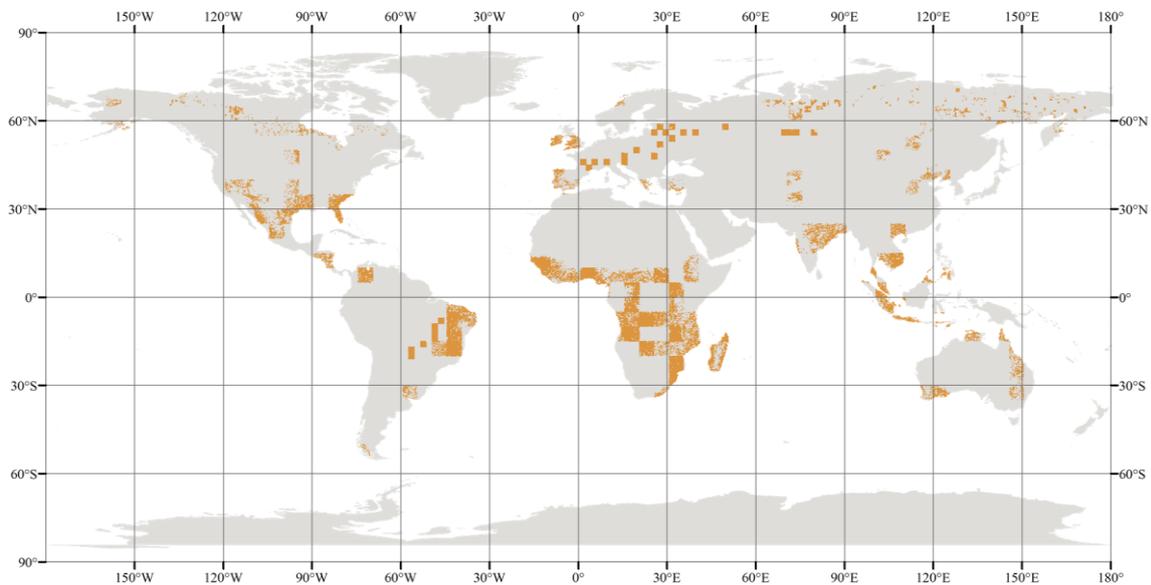

Figure 6 The distribution of training samples.

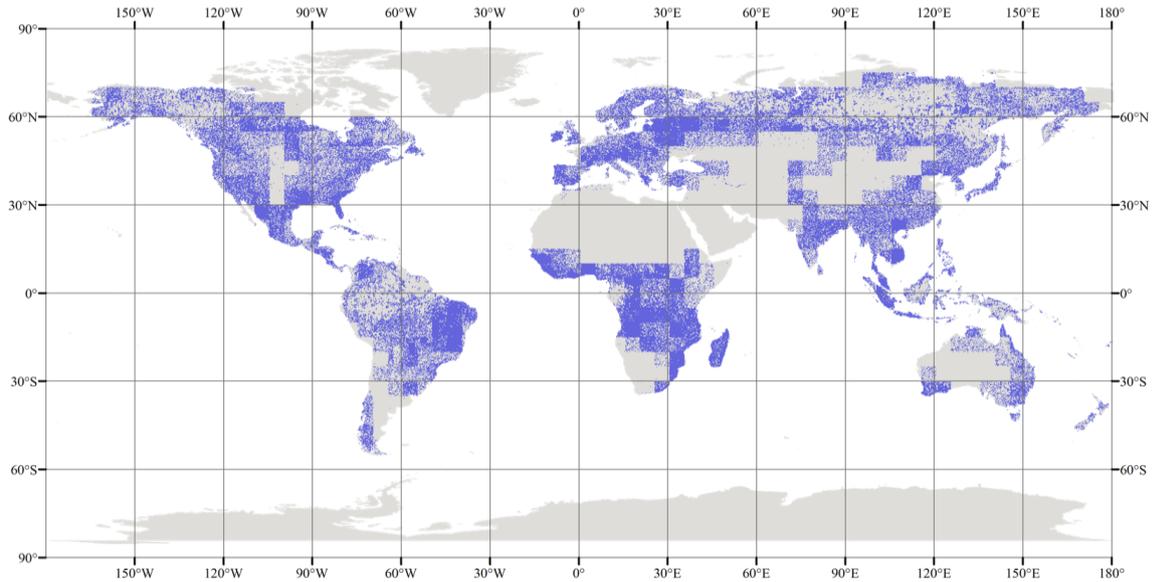

Figure 7 The distribution of validation samples.

## 3.2 Classification with Different Sources of Training Samples

In this section, we try to generate a more accurate classification map from the view of training sample set, which is the most important component of classification process. How to fully utilize previous GFC products to generate more confident training samples with minimum manpower is the key problem needed to be solved in this section. We can find that the samples consistently voted by previous GFC products have high confidence belong to voted category. Thus, we will evaluate the accuracy of samples under different sample certainty and investigate their potential to be used as training samples. Secondly, different sampling strategies from local and global scale will be discussed in this section.

### 3.2.1 Sample Certainty

If a sample is voted by most of GFC products to one category, it can be referred as "certain samples". On the contrary, a sample with controversial voted results can be referred as "uncertain samples". In the case of insufficient training samples, the "certain samples" can be directly used as training samples input to classifier, which can be called as "free training samples". In this section, we carry out thorough analysis of the accuracy and training availability of these "free training samples".

We regard the sample whose forest vote by five GFC products equals 0 or 1 as non-forest, and regard the sample whose forest vote by five GFC products equals 2, 3, 4 or 5 as forest.

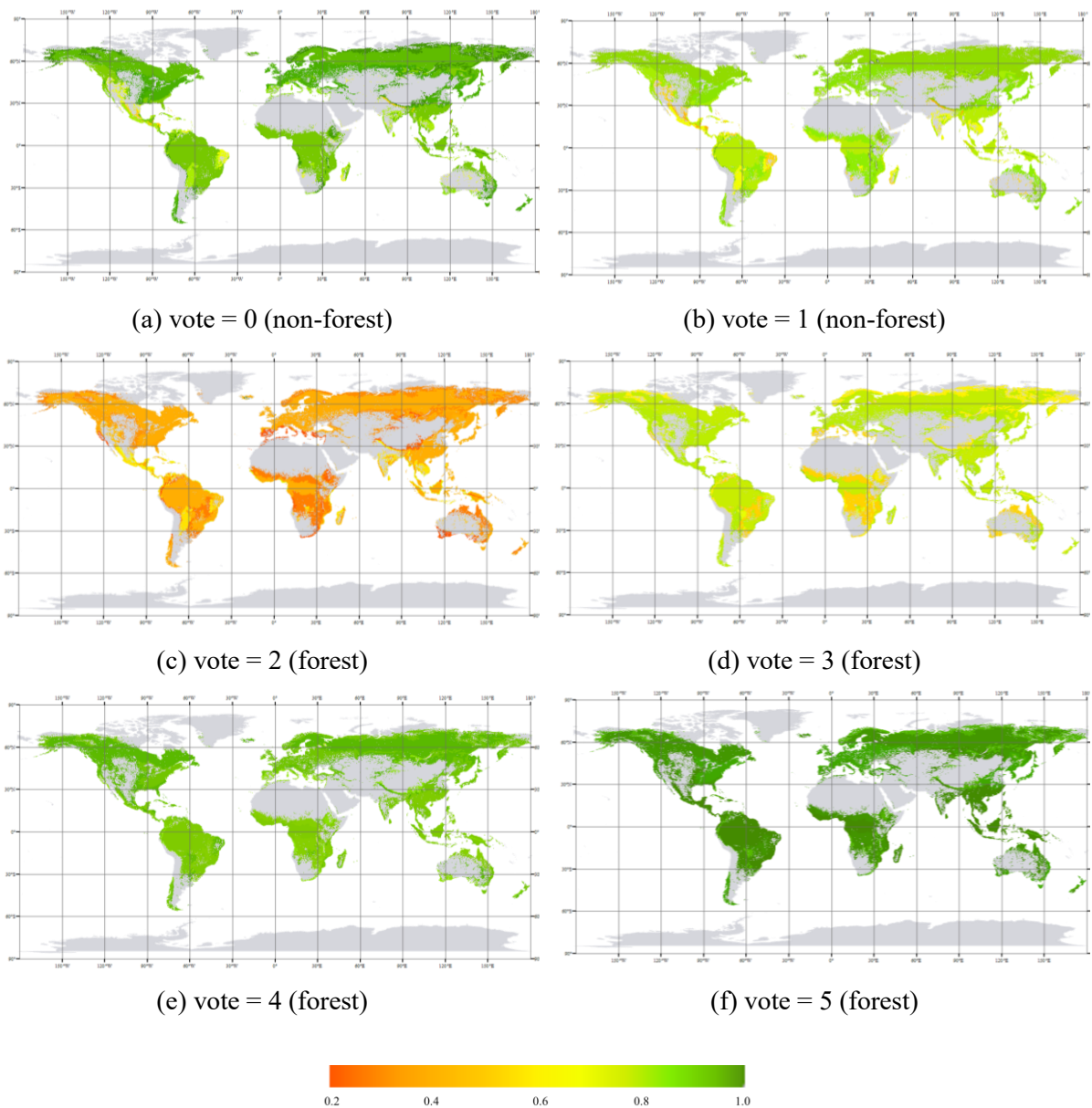

(a) vote = 0 (non-forest)   (b) vote = 1 (non-forest)
(c) vote = 2 (forest)       (d) vote = 3 (forest)
(e) vote = 4 (forest)       (f) vote = 5 (forest)

Figure 8 Type certainty of each number of votes in each ecoregion.

The type certainty (TC) of each number of votes in each ecoregion is calculated. The type certainty is calculated as follows:

$$TC_{class,vote} = \frac{COUNT(P_{vote}^{class})}{COUNT(P_{vote})} \quad (4)$$

$$class \in \{forest, non-forest\} \quad (5)$$

where $P_{vote}$ is the point whose value of agreement map (Figure 1) equals *vote*, $P_{vote}^{class}$ is the point whose value of agreement map (Figure 1) equals *vote* and is labeled to *class*.

The basic unit of calculation and display is ecoregion: only the regions whose forest votes of five sets of products are greater than 0 are displayed in Figure 8. Figure 8 (a) shows that in a few ecoregions the type certainty of 0 sets is a little low. This means that the actual type of many sample points of forest votes being 0 is forest in these ecoregions, which indicates that there are many forests omitted in these ecoregions. Figure 8 (b) shows the proportion of the actual type being non-forest to the sample points whose forest votes of five sets of products equal to 1 in each ecoregion. Figure 8 (c-f) display the proportion of the actual forest samples to the samples whose forest votes equal to 2,3,4 and 5 respectively. From Figure 9 it can be found that the median $TC_{non-forest,1}$ is about 82%, and the median $TC_{forest,2}$, $TC_{forest,3}$, $TC_{forest,4}$ and $TC_{forest,5}$ is about 43%, 79%, 92% and 97%, respectively. Generally, sampling points whose forest votes of five sets of products equal to 4 and 5 have high type certainty, which means that samples with votes being 4 and 5 in these regions can be used as training forest samples directly. Sample points whose forest votes of five sets of products equal to 0 have high type certainty in most ecoregions, so the points with votes being 0 in these ecoregions can be used as training non-forest samples directly, and the labeled non-forest sample points will be used as training samples in other ecoregions.

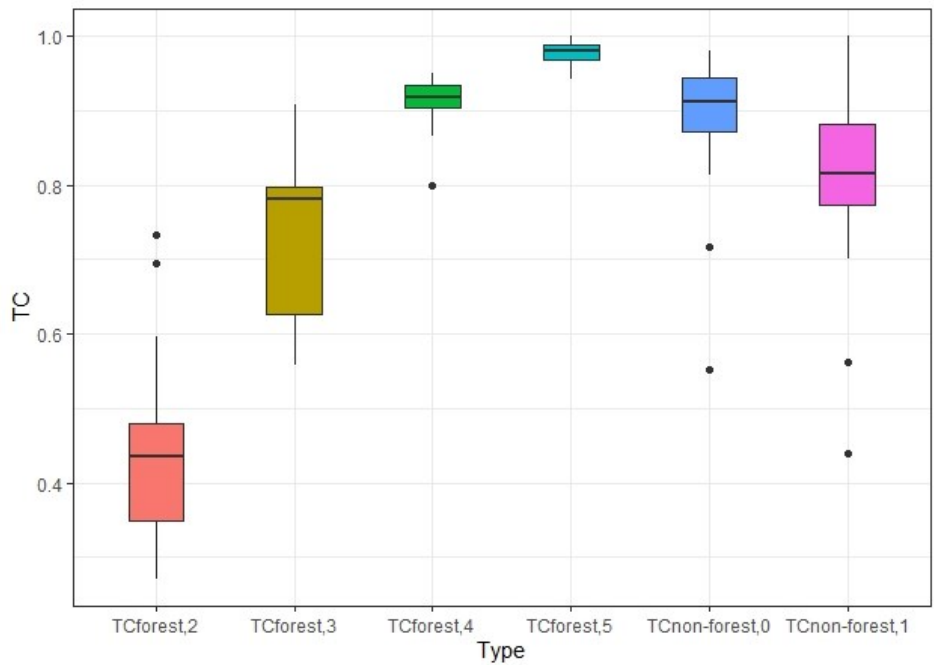

Figure 9 Box plots for each type certainty (TC).

Figure 10 shows some samples inconsistent with forest products and labels, which means that there are some forests in the regions whose forest votes equal to 0 and 1, and some non-forests in the regions whose forest votes equal to 4 and 5. Figure 10 (a) shows some samples whose forest votes equal to 0 and 1 but marked as forest, and Figure 10 (b) shows some samples whose forest votes equal to 4 and 5 but marked as non-forest. It can be found that the samples whose forest votes equal to 0 and 1 but marked as forest are mainly sparse forests and plantations, while the samples whose forest votes equal to 4 and 5 but marked as non-forest are mainly shrublands.

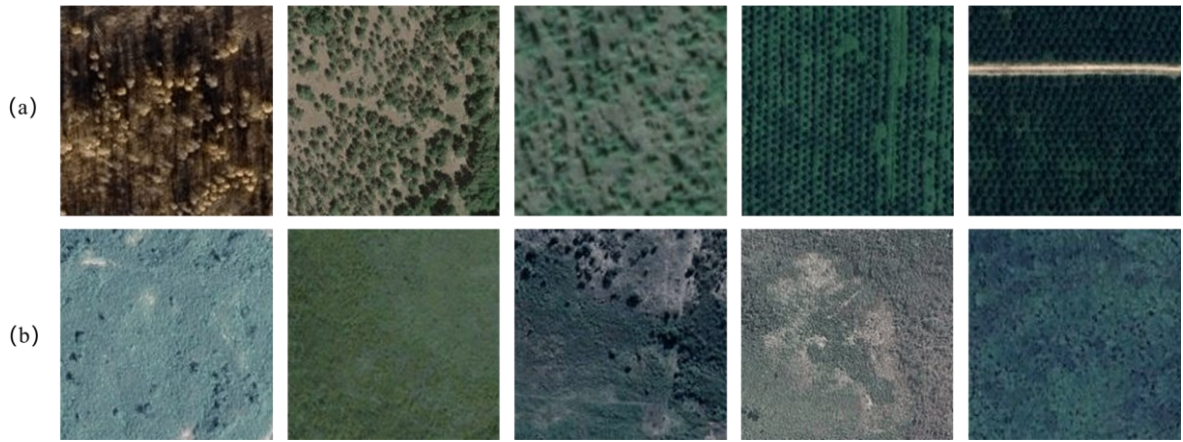

Figure 10 Some samples inconsistent with forest products and labels.

### 3.2.2 Feature Extraction

In this study, the 2020 time-series of Landsat SR imagery and corresponding spectral indexes, including NDVI (Normalized Difference Vegetation index) (Tucker, 1979) and NDWI (Normalized Difference Water Index) (Xu, 2006) are considered. The spectral features include median value of 2020 yearly bands of blue, green, red, near infrared, shortwave infrared 1, shortwave infrared 2 surface reflectance and NDVI calculated from median near infrared and red values, NDWI calculated from median green and near infrared values. Except for these Landsat-based metrics, the topographical variable of slope, derived from the DEM datasets (Yamazaki et al., 2017), were also added. All these features are fed to random forest classifier.

To ensure the richness of sample features, we used NDVI as the stratification attribute and divided it into 10 groups with the step of 0.2. 500 sample points were selected in each stratum to input to the random forest classifier. Although random forest classification was performed in a 5° grid unit, to ensure adequacy of samples, we applied sampling not only within the 5° grid, but also additionally in the tnc_terr_ecoregions (Data sources: The Nature Conservancy, vector digital data available at http://maps.tnc.org/gis_data.html ). It will be demonstrated that samples from the same ecoregion helps improving classification accuracy in section IV.

### 3.2.3 Sampling Strategies

Having accumulated a large high quantity of the labeled samples as mentioned in section 3.1, a discussion on the sampling strategy to be adopted is necessary for the accuracy of the classifier. We conducted experiments to discuss training data allocation strategies that utilize contexts at three different levels, i.e., within the grid (Local) which is also to be classified; within the ecoregion where the grid locates in (Ecoregion); and within the global region (Global). There are five strategies needed to be tested (see Table 6) to explore which one can achieve the best performance.

Table 6 Different training data allocation strategies.

| Strategies | Contexts |
| --- | --- |
| Strategy 1 | Global |
| Strategy 2 | Local |
| Strategy 3 | Global + Local |
| Strategy 4 | Ecoregions +Local |
| Strategy 5 | Ecoregions |

### 3.2.4 Accuracy Assessment for Our and Previous GFC Products

Accuracy metrics include UA, PA, OA, Kappa. The total number of correct samples of forest divided by the total number of samples of forest derived from the reference data, indicates the probability of a reference sample being correctly classified and is really a measure of omission error. This accuracy measure is often called "producer's accuracy (PA)". On the other hand, the total number of correct samples of forest divided by the total number of samples that are classified to forest, is called "user's accuracy" (UA). "Overall accuracy" (OA) is computed by dividing the total correct samples by the total number of samples in the error matrix. Kappa analysis provides tests of significance. With this technique, it is possible to test whether individual land cover map generated from remotely sensed data is much better than the map generated by randomly assigned area labels (Congalton, 2001). These four metrics are defined as follows with an example error matrix (Table 7).

Table 7 Example error matrix.

| | | True label | |
| --- | --- | --- | --- |
| | | forest | non-forest |
| Predicted label | forest | *NUMa* | *NUMc* |
| | non-forest | *NUMb* | *NUMd* |

$$UA = \frac{NUMa}{NUMa + NUMb} \tag{6}$$

$$PA = \frac{NUMa}{NUMa + NUMc} \tag{7}$$

$$OA = \frac{NUMa + NUMd}{NUMa + NUMb + NUMc + NUMd} \tag{8}$$

$$Kappa = \frac{OA - Pe}{1 - Pe} \tag{9}$$

$$Pe = \frac{(NUMa + NUMb) * (NUMa + NUMc) + (NUMb + NUMd) * (NUMc + NUMd)}{(NUMa + NUMb + NUMc + NUMd)^2} \tag{10}$$

# IV. Experimental result

In this section, we first verify the effectiveness of certain samples and manually annotated samples in improving classification accuracy, then discuss which sampling methods would better exploit the advantages of training samples, and finally provide a comparative analysis of five previous global forest products as well as our forest product.

## 4.1 Effectiveness of Free Samples and Manually Labeled samples

In the previous section, the manual inspection of several sets of certain samples confirmed that most of the certain samples are correct and can be used directly in our classification process. Therefore, larger labor cost is invested in the labeling of uncertain samples. This section discusses whether the feeding of free samples and labeled uncertain samples into classifier can significantly increase the classification accuracy.

As mentioned in section 3.2.1, pixels with votes of 4 or 5 have high certainty to be forest and those with votes of 0 are certain to be non-forest. Therefore, forest layers of 5 products are used to construct part of our samples as certain samples. Pixels with vote values of 4 or 5 are considered as forest samples, and pixels with vote values of 0 are considered as non-forest samples. Since these certain samples are from regions with high agreement where most of the products can be classified to the same category and may have more discriminative spectral features, these certain samples may lack necessary information to classify the uncertain regions. Therefore, we randomly sample pixel points with values of 2 or 3 as uncertain samples and manually label whether they are forests or not.

Three comparison experiments are designed to explore the role of certain samples and uncertain samples in classification. We only used 5000 certain samples as input of random forest classifier in the CSRF (Certain Sample Random Forest) strategy. Besides, about 1000 manually labeled uncertain samples were put into the classifier in USRF (Uncertain Sample Random Forest) strategy. Certain samples and uncertain samples are combined as input in CUSRF (Certain and Uncertain Sample Random Forest) strategy. Although samples in USRF are less than that in other strategies, they still show great effect on accuracy enhancement. And sample sizes of CSRF and CUSRF are constrained to be equal for fair comparison. In ecoregions where the uncertain grids are located, the classification accuracy of CSRF using only certain samples is the lowest, while the classification accuracy of CUSRF using uncertain samples and certain samples is the highest in most of the ecoregions, and the accuracy of USRF using only uncertain samples is generally higher than CSRF (Figure 11). In detail display (Figure 12), although these three classification results show spatially similar and can depict the distribution and shape of forest, results of CUSRF achieve best in the local accuracy assessment.

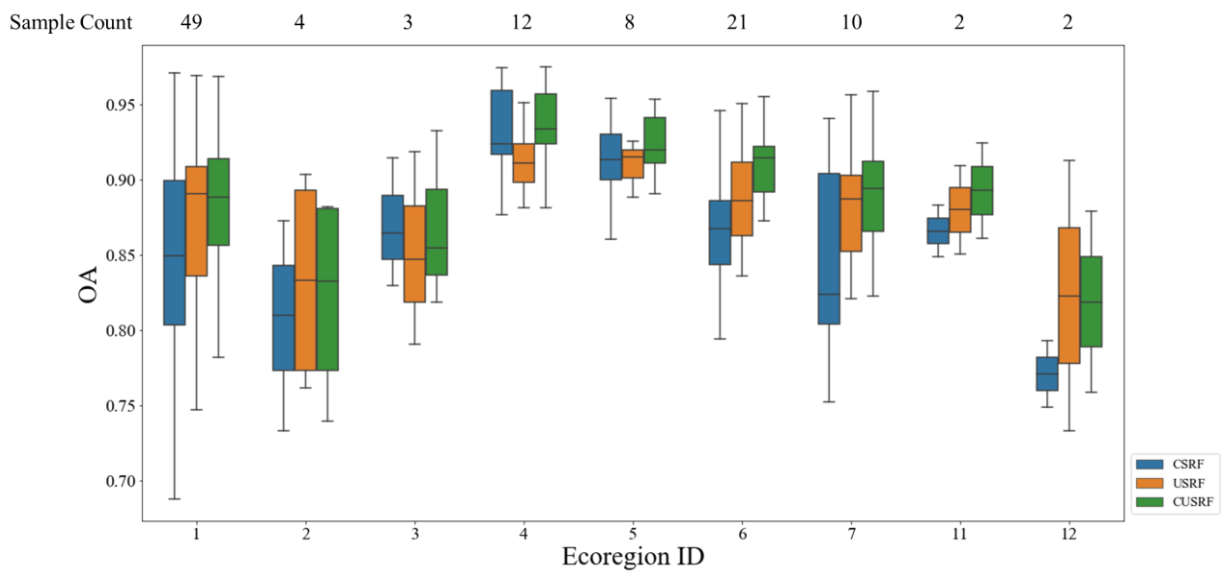

Figure 11 Boxplot of accuracy of classification with different inputs.

**CSRF**

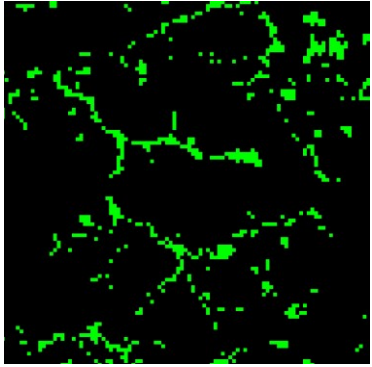 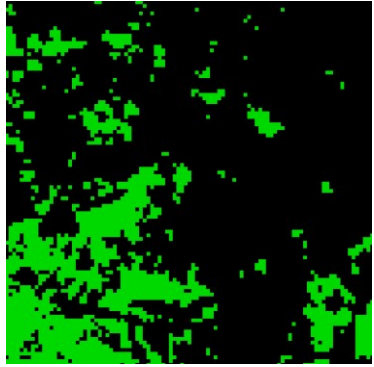 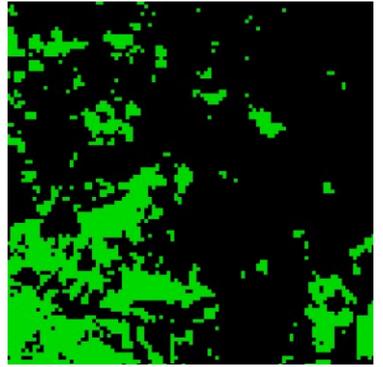

| | | | |
|---|---|---|---|
| OA | 0.8604 | 0.8919 | 0.8154 |
| Kappa | 0.2609 | 0.6524 | 0.5185 |

**USRF**

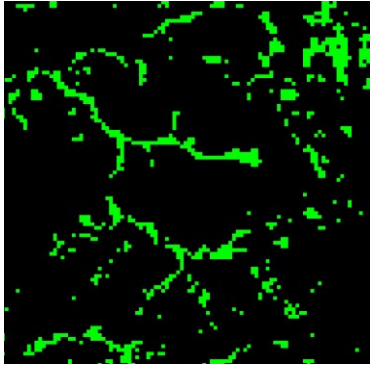 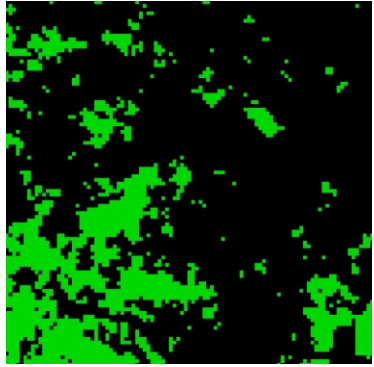 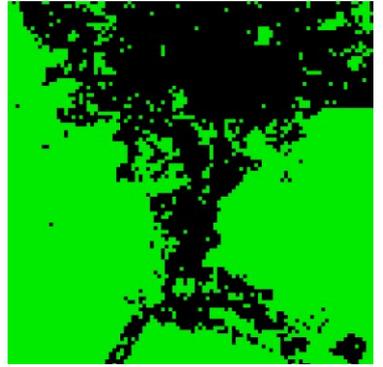

| | | | |
|---|---|---|---|
| OA | 0.8598 | 0.8938 | 0.8531 |
| Kappa | **0.2993** | 0.6614 | **0.6790** |

**CUSRF**

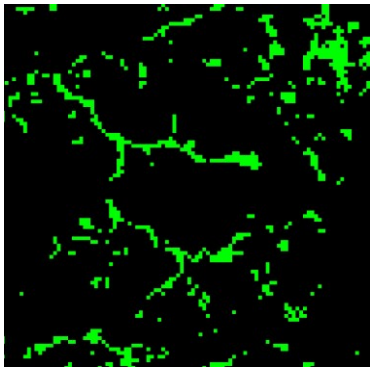 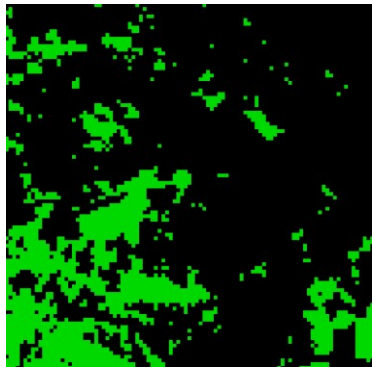 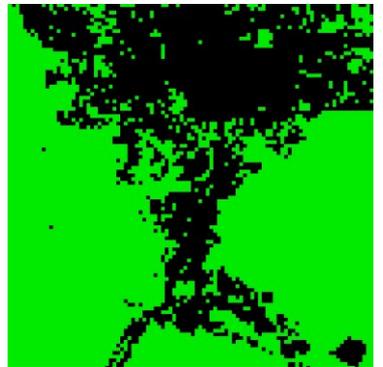

| | | | |
|---|---|---|---|
| OA | **0.8624** | **0.9025** | **0.8556** |
| Kappa | 0.2925 | **0.6818** | 0.6782 |

**Ground Truth**

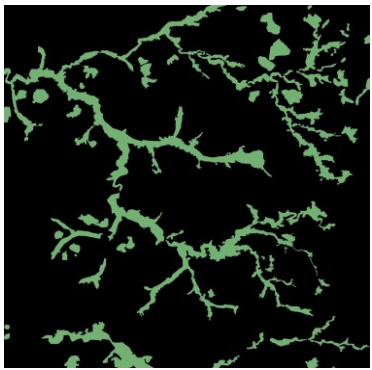 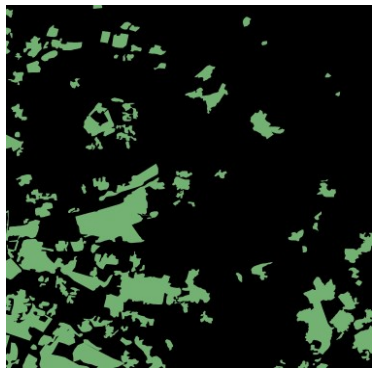 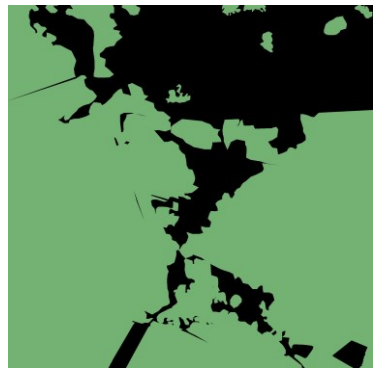

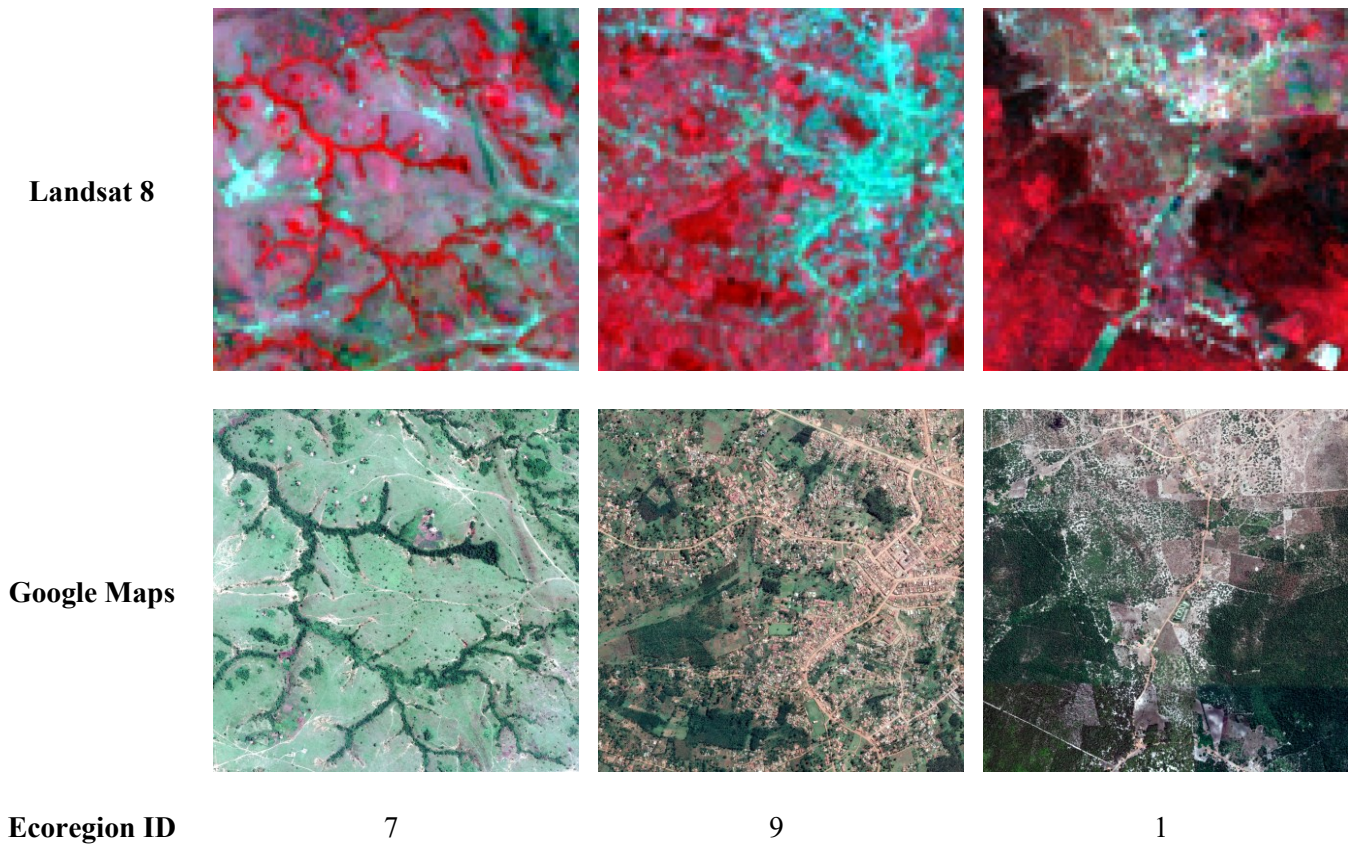

| | | | |
|---|---|---|---|
| **Landsat 8** | | | |
| **Google Maps** | | | |
| **Ecoregion ID** | 7 | 9 | 1 |

Figure 12 Classification results of different strategies and forest products of 30m resolution. (Ecoregion ID and corresponding ecoregion: ID 1 is Tropical and Subtropical Moist Broadleaf Forests, ID 7 is Tropical and Subtropical Grasslands, Savannas and Shrublands, ID 9 is Flooded Grasslands and Savannas)

The five products and the three classification methods are compared. As not all products are available in the 2020 version, we could only select their versions from the most recent years for classification as well as comparison, assuming that the area change from 2017 to 2020 is small. In the uncertain regions (Table 8), we find that the overall accuracy of CSRF exceeds other products except FROM_GLC30 2017. The accuracy of USRF exceeds other products, indicating that manually labeled samples have a positive effect on the classification. The highest accuracy is achieved by CUSRF, which demonstrates the combination of certain samples and uncertain samples help achieve the best accuracy.

Table 8 Accuracy of global forest products and classification strategies.

| Product | PA | UA | OA | Kappa |
|---|---|---|---|---|
| FROM_GLC30 2017 | 0.7508 | **0.9361** | 0.8469 | 0.6948 |
| GFC30 2018 | 0.7902 | 0.8215 | 0.8056 | 0.6113 |
| GLADForest 2019 | 0.7832 | 0.8356 | 0.8110 | 0.6223 |
| GLC_FCS30-2020 | 0.8414 | 0.7405 | 0.7689 | 0.5363 |
| GlobeLand30 2020 | 0.7204 | 0.8041 | 0.7680 | 0.5368 |
| CSRF | **0.9084** | 0.7972 | 0.8355 | 0.6700 |
| USRF | 0.8432 | 0.8726 | 0.8573 | 0.7148 |
| CUSRF | 0.8626 | 0.8881 | **0.8746** | **0.7492** |

## 4.2 Performances of Different Sampling Strategies

To select the best of the five sampling strategies mentioned in 3.2.3, we classified 12 grids distributed in different ecoregions, including Tropical and Subtropical Moist Broadleaf Forests, Temperate Broadleaf and Mixed Forests, Boreal Forests/Taiga, Tropical and Subtropical Grasslands and Savannas and Shrublands (see Figure 13). The results of each strategy were compared hence creating the overall accuracy and Kappa (see Table 9 and Figure 14). The overall accuracies of all grids of the five strategies are 0.894 (Strategy 1), 0.913 (Strategy 2), 0.916 (Strategy 3), 0.928 (Strategy 4), 0.888 (Strategy 5), respectively. The fourth strategy (Ecoregion + Local) achieves higher accuracy than other strategies. It indicates that samples from the same ecoregion provide similar forest information, which can solve the problem of insufficient samples in grids. Samples from the ecoregion where the grid locates provide better information than that from the global area. When only one specific sampling scale is considered (i.e., Strategy1, Strategy2, Strategy5), the training data may lose part of useful local feature or global feature. Moreover, dispersed distributed samples may solve the problem of local overfitting. It indicates that combining samples at local and ecoregion scale is beneficial to improve the classification accuracy.

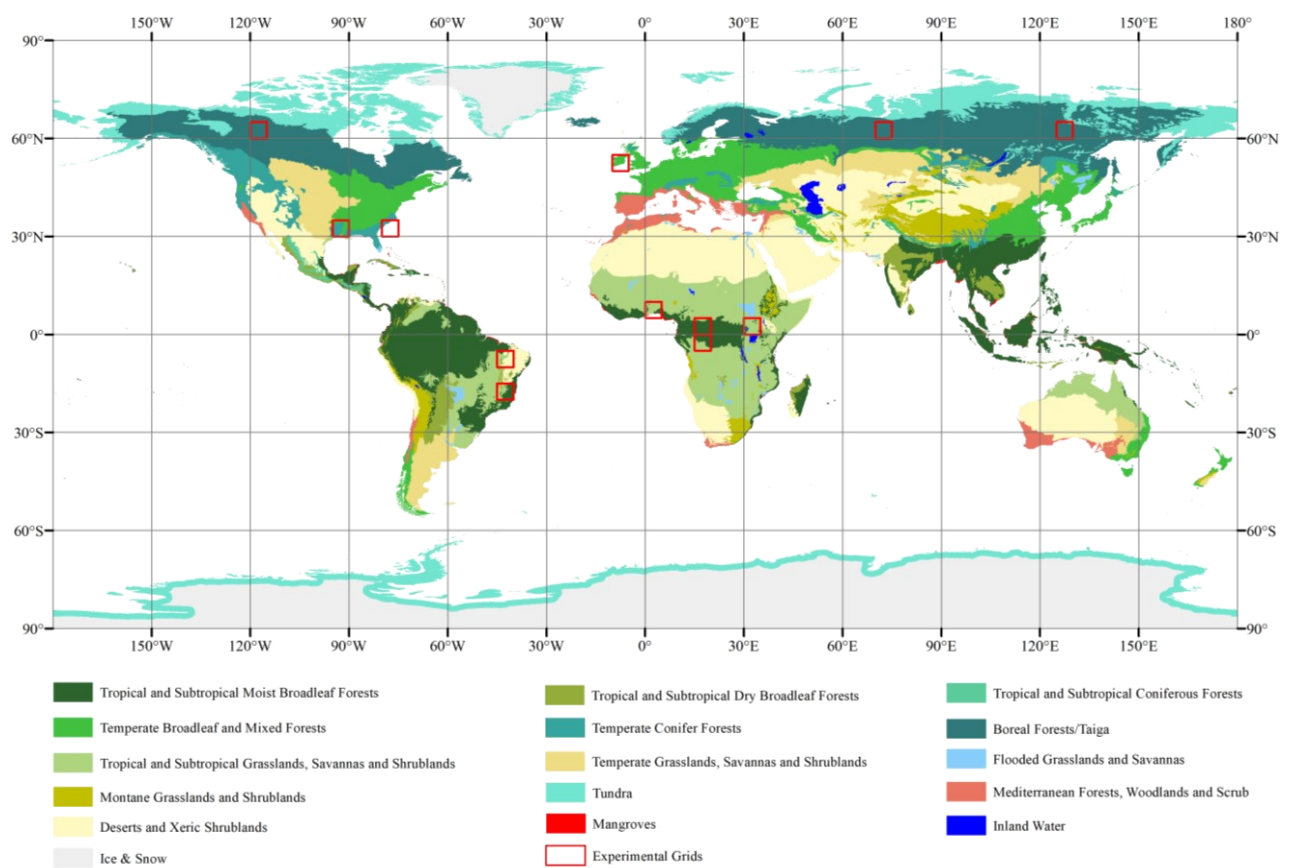

Figure 13 The distribution of 12 experimental grids in the ecoregions.

Table 9 Accuracy of different training data allocation strategies.

| | Strategy 1 | | Strategy 2 | | Strategy 3 | | Strategy 4 | | Strategy 5 | |
|---|---|---|---|---|---|---|---|---|---|---|
| | OA | *Kappa* | OA | *Kappa* | OA | *Kappa* | OA | *Kappa* | OA | *Kappa* |
| Grid 1 | 0.9209 | 0.8419 | 0.9007 | 0.8015 | 0.9173 | 0.8346 | **0.9274** | **0.8548** | 0.9195 | 0.8391 |
| Grid 2 | 0.8377 | 0.6753 | 0.9247 | 0.8494 | 0.9137 | 0.8274 | **0.9461** | **0.8923** | 0.8372 | 0.6743 |
| Grid 3 | 0.8117 | 0.6234 | 0.8194 | 0.6389 | 0.8219 | 0.6537 | **0.8227** | 0.6454 | 0.8139 | **0.6687** |
| Grid 4 | 0.9110 | 0.8219 | 0.8894 | 0.7788 | 0.9146 | 0.8291 | **0.9310** | **0.8620** | 0.9137 | 0.8273 |
| Grid 5 | 0.8901 | 0.7802 | 0.9038 | 0.8077 | 0.8791 | 0.7582 | **0.9359** | **0.8718** | 0.8919 | 0.7839 |
| Grid 6 | 0.8969 | 0.7938 | 0.9396 | 0.8792 | 0.9396 | 0.8792 | **0.9492** | **0.8984** | 0.9300 | 0.8601 |
| Grid 7 | 0.9628 | 0.9256 | 0.9781 | 0.9561 | 0.9759 | 0.9519 | **0.9807** | **0.9614** | 0.9199 | 0.8398 |
| Grid 8 | 0.9166 | 0.8333 | 0.9117 | 0.8233 | 0.9248 | 0.8496 | **0.9451** | **0.8902** | 0.8934 | 0.7867 |
| Grid 9 | 0.8975 | 0.7950 | 0.9065 | 0.8130 | 0.9054 | 0.8109 | **0.9069** | **0.8134** | 0.8997 | 0.7994 |
| Grid 10 | 0.9628 | 0.9256 | 0.9605 | 0.9201 | 0.9613 | 0.9230 | 0.9615 | 0.9231 | **0.9641** | **0.9282** |
| Grid 11 | 0.9717 | 0.9434 | 0.9695 | 0.9389 | 0.9683 | 0.9367 | **0.9734** | **0.9468** | 0.9666 | 0.9333 |
| Grid 12 | **0.9389** | **0.8778** | 0.9359 | 0.8717 | 0.9349 | 0.8697 | 0.9369 | 0.8737 | 0.9259 | 0.8517 |

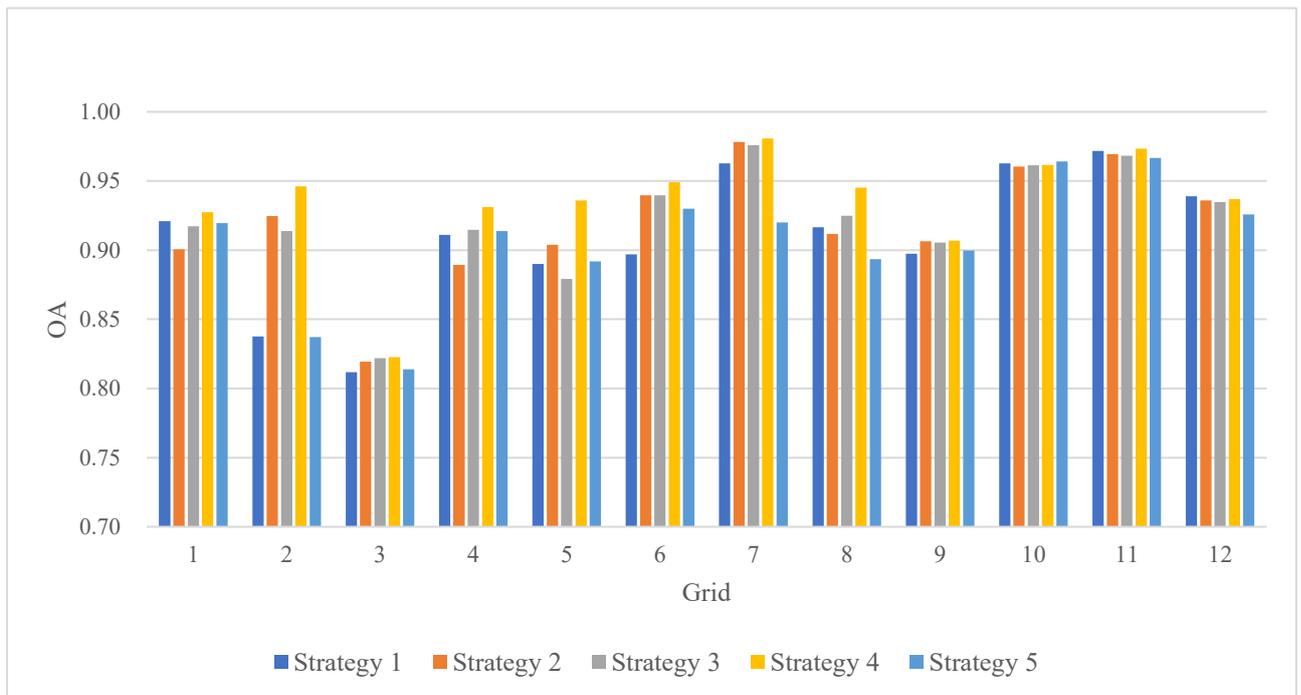

Figure 14 Overall accuracy of different training data allocation strategies.

## 4.3 Global Forest Classification Result

### 4.3.1 Global Forest Classification Maps

Figure 15 shows the forest proportion in each 0.5°×0.5° geographical grid of six global 30m forest products. All global 30m forest products can accurately capture the main distribution pattern of forest. For example, it can be found from six forest products that forest is mainly distributed in tropical rainforest areas such as the Amazon rainforest, Indian–Malay rainforests and African rainforests. However, there are also some differences in the distribution of forest among these forest products. In Africa, GLADForest 2019 and GLC_FCS30-2020 report higher forest densities than other forest maps, while GlobeLand30 2020 and FROM_GLC30 2017 have less forest densities. All products show high forest density in South America, except GlobeLand30 2020. In some high-latitude areas such as Europe and North America, GlobeLand30 2020 and GLC_FCS30-2020 have lower forest densities than other forest maps. GlobeForest2020 has highest forest intensity in Asia, Europe and eastern Australia. We calculated the area of global forest of the six products, and found that GlobeForest2020 has the largest forest area with 4.65E+07 km², followed by GLC_FCS30-2020 with 4.33E+07 km². FROM_GLC30 2017 and GLADForest 2019 have similar forest area. GFC30 2018 has minimal forest area with 2.85E+07 km² (see Figure 16).

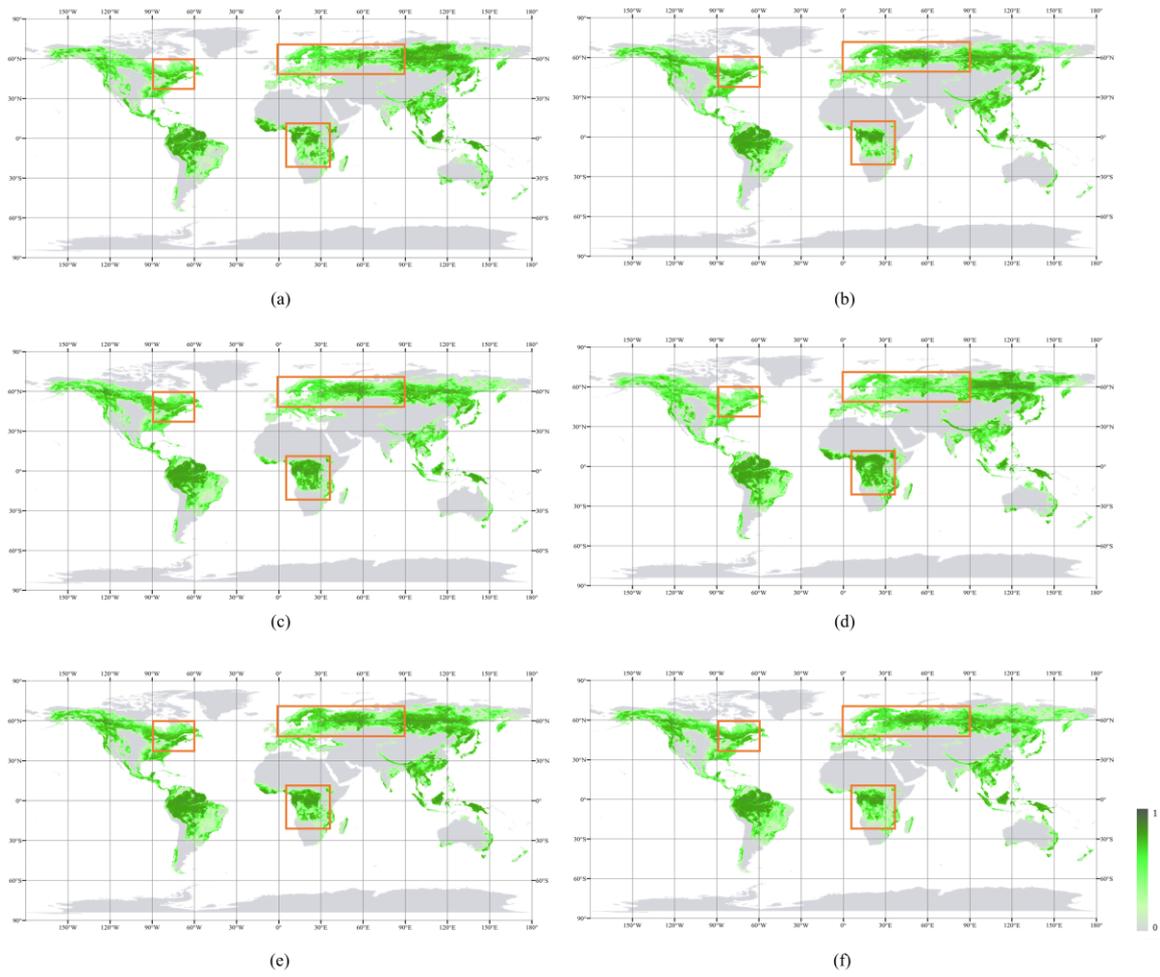

Figure 15 Forest maps after aggregation to a resolution of 0.5°. (a) GlobeLand30 2020. (b) FROM_GLC30 2017. (c) GLADForest 2019. (d) FCS_GLC30-2020. (e) GlobeForest2020. (f) GFC30 2020.

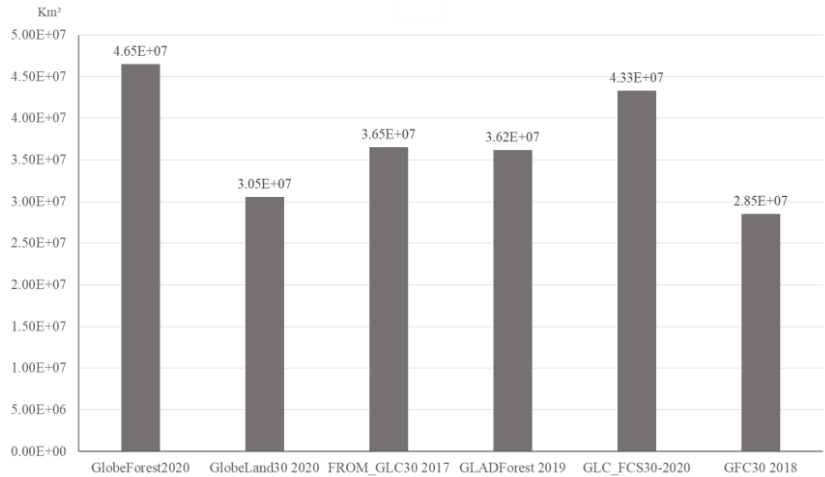

Figure 16 Area Statistics of all forest products.

## 4.3.2 Local View Display

The four ecoregions with the highest uncertainty were selected for comparison (Figure 17-Figure 20). In Tropical and Subtropical Dry Broadleaf Forests (Figure 17) and Tropical and Subtropical Coniferous Forests (Figure 18), the results for FROM_GLC30 2017, GFC30 2018, GlobeLand30 2020 and GlobeForest2020 are very consistent, while the classification results for GLC_FCS30-2020 and GLADForest 2019 appear over- or under-classified in forest. As for Flooded Grasslands and Savannas (Figure 19), results of FROM_GLC30 2017, GlobeLand30 2020 and GlabeForest2020 are similar, while the results of the remaining three products vary considerably. These six products have very similar classification profiles for the Mediterranean Forests, Woodlands and Scrub (Figure 20), with only minor differences in detail. GlobeForest2020 achieves the first or second highest accuracy in these four local areas.

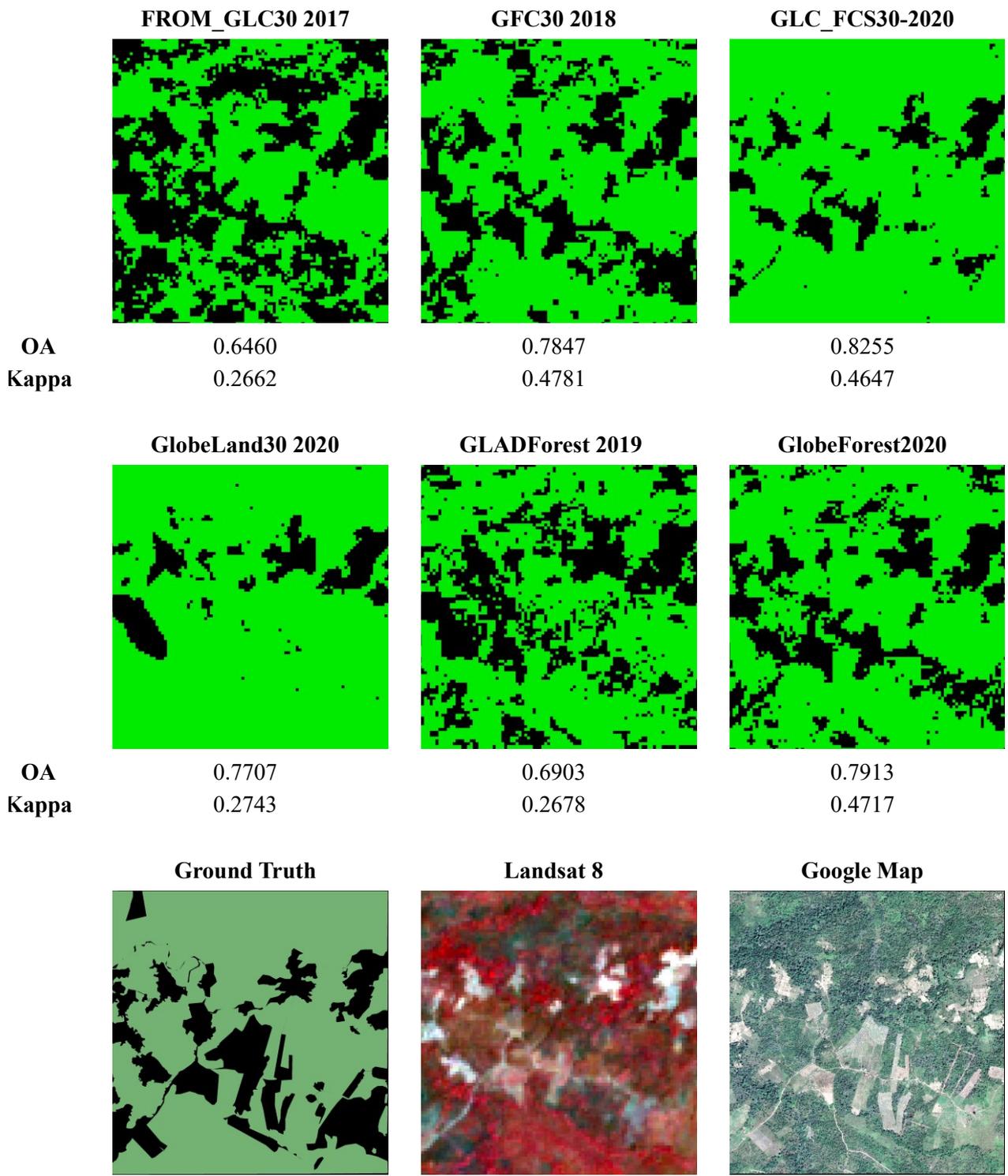

Figure 17 Local view of forest maps in ecoregion of Tropical and Subtropical Dry Broadleaf Forests.

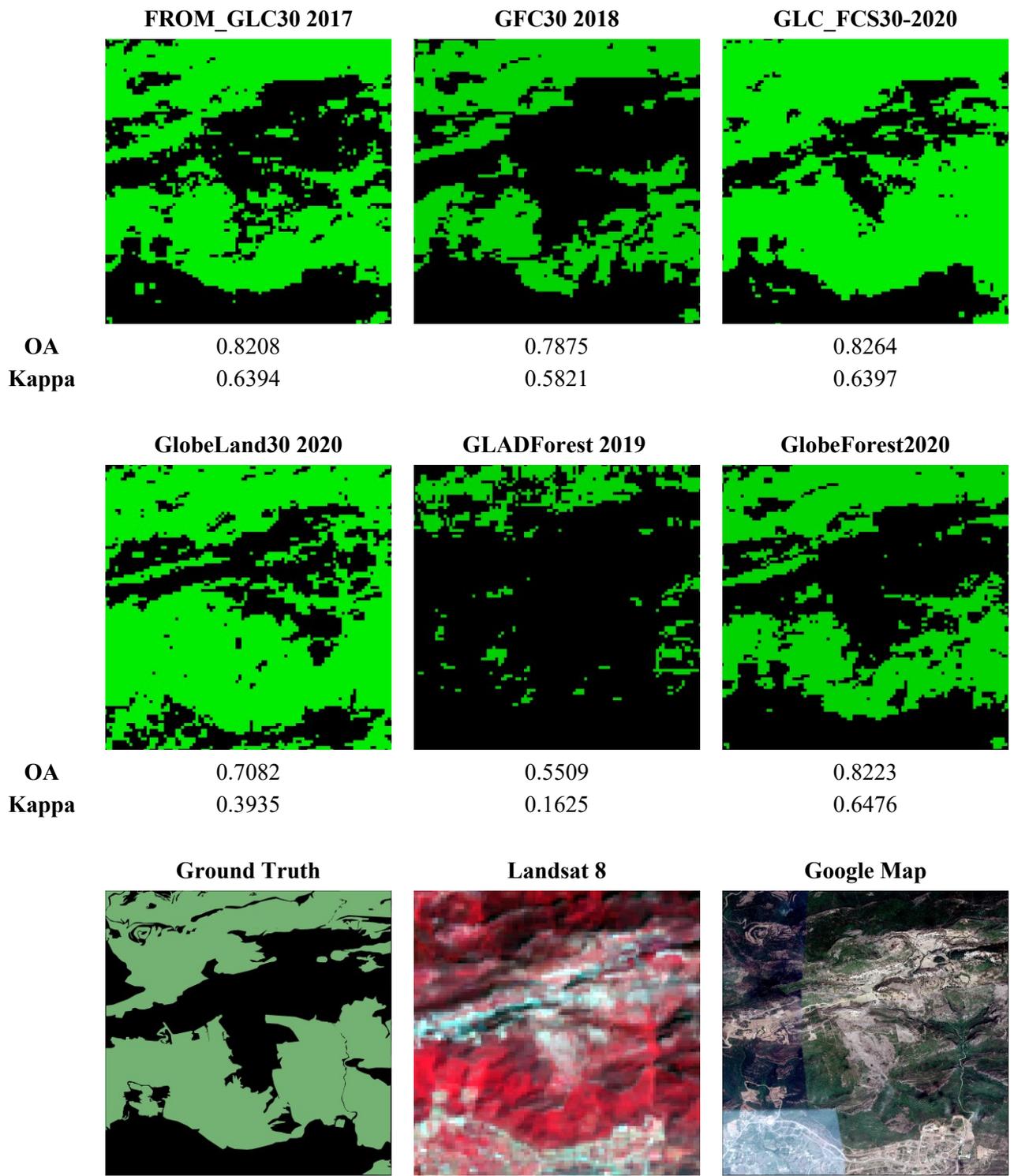

Figure 18 Local view of forest maps in ecoregion of Tropical and Subtropical Coniferous Forests.

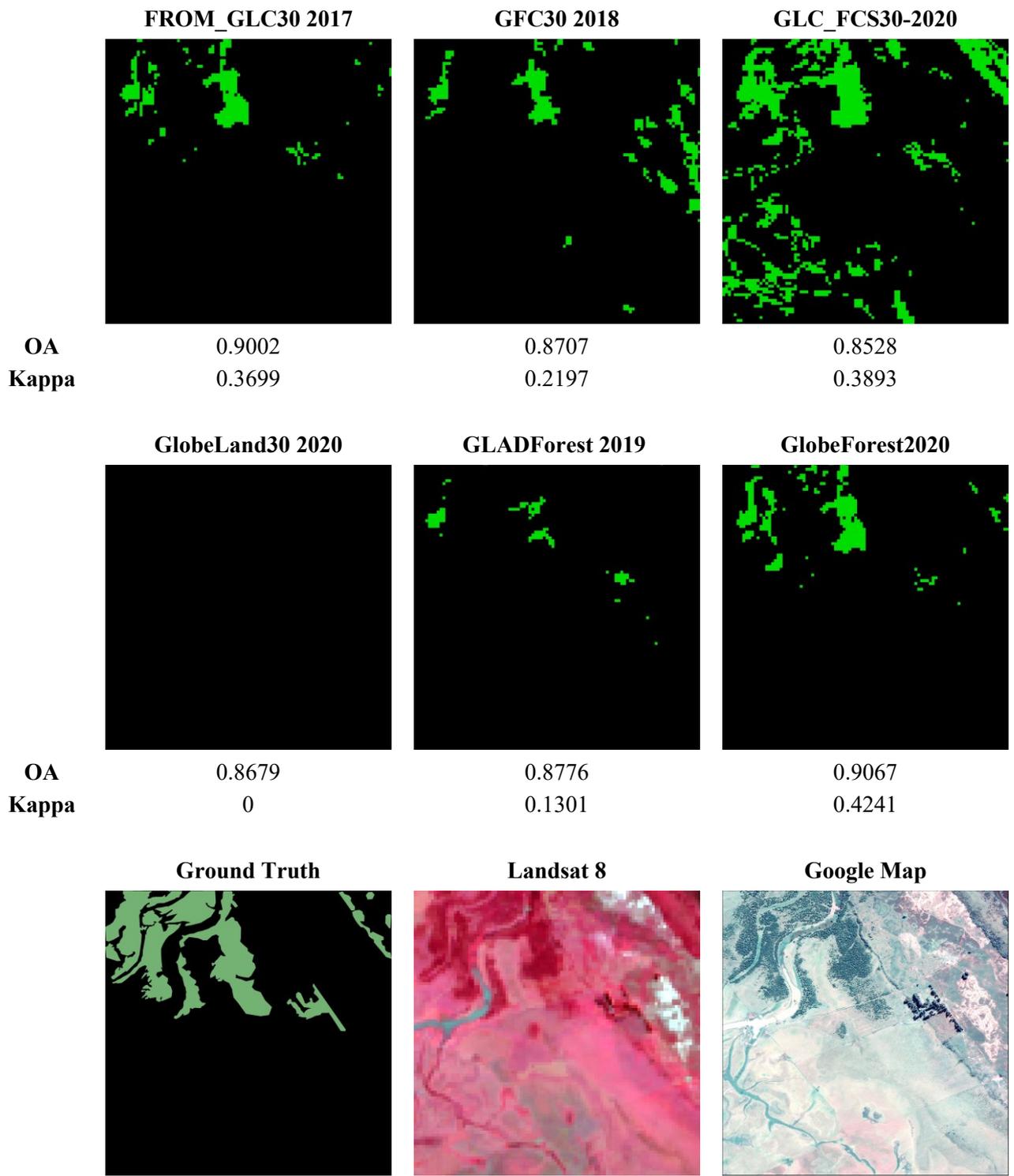

Figure 19 Local view of forest maps in ecoregion of Flooded Grasslands and Savannas.

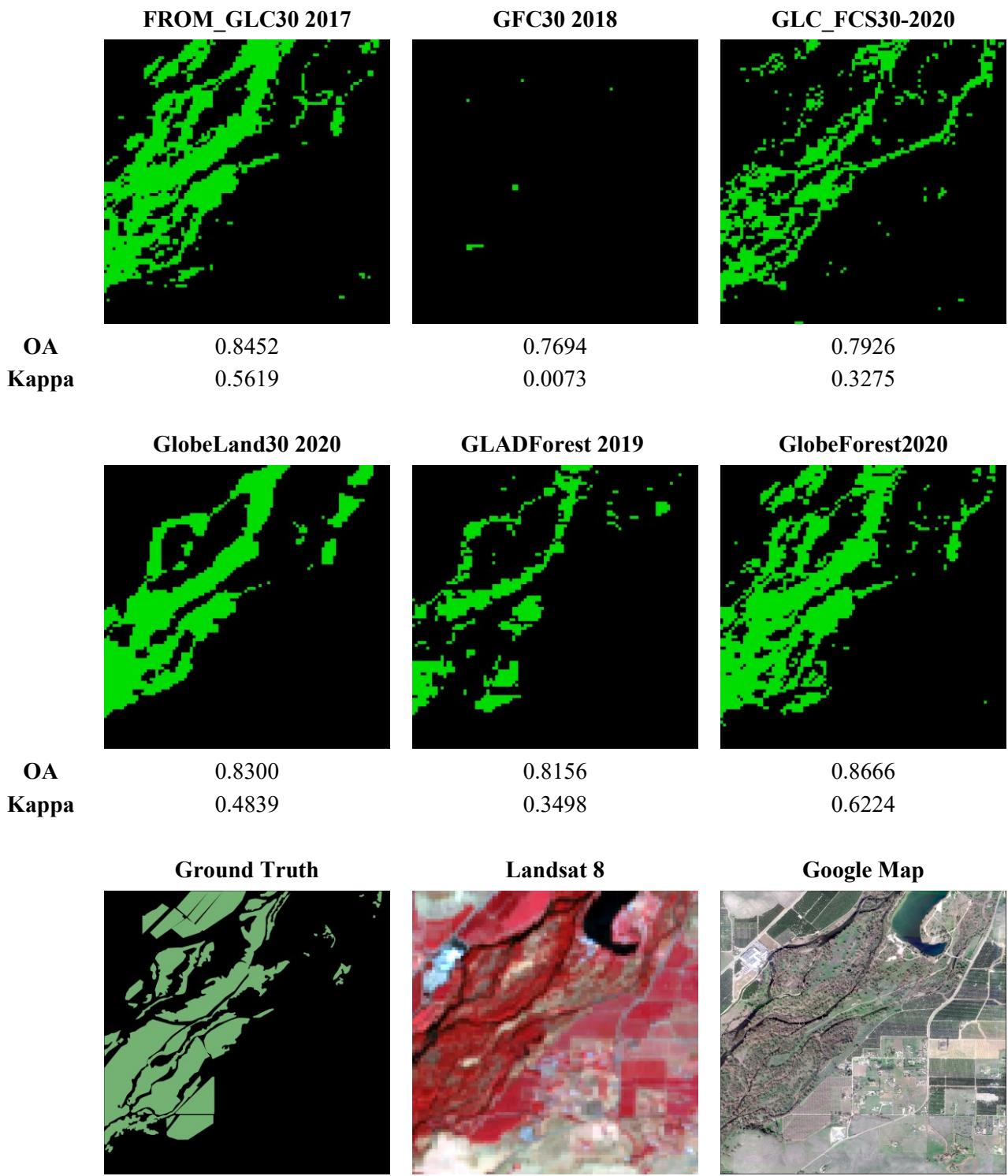

Figure 20 Local view of forest maps in ecoregion of Mediterranean Forests, Woodlands and Scrub.

### 4.3.3 Accuracy Assessment

Accuracies in terms of four metrics are calculated in 5° grids, and box plots are produced for each product for all grids on a global scale (Figure 21) or within ecoregions (Figure 22). UA of GlobeForest2020 is slightly less than the highest of other products by 1.14%. PA of GlobeForest2020 exceeds the highest of others by 2.31%, and it also achieves the best OA and Kappa. When aggregating accuracies of 6 products in unit of ecoregions, GLADForest 2019, FROM_GLC30 2017 and GlobeForest2020 perform better in terms of UA, which means that the predicted forest has higher probability of being a forest. GLC_FCS30-2020, FROM_GLC30 2017 and GlobeForest2020 perform better in terms of PA, namely larger proportion of forests pixels are correctly predicted. In ecoregions of Tropical and Subtropical Dry Broadleaf Forests, Tropical and Subtropical Coniferous Forests, Tropical and Subtropical Grasslands, Savannas and Shrublands, Mediterranean Forests and Woodlands and Scrub (i.e., with Ecoregion ID of 2, 3, 7, 12), OA and Kappa of 6 products are relatively low. This may be attributable to the high uncertainty of these four ecoregions as mentioned in Table 2. In these regions, although FROM_GLC30 2017 and GlobeForest2020 perform best in terms of OA and Kappa, GlobeForest2020 has shorter boxes, indicating that it has a smaller variation in accuracies across all grids and performs more steadily.

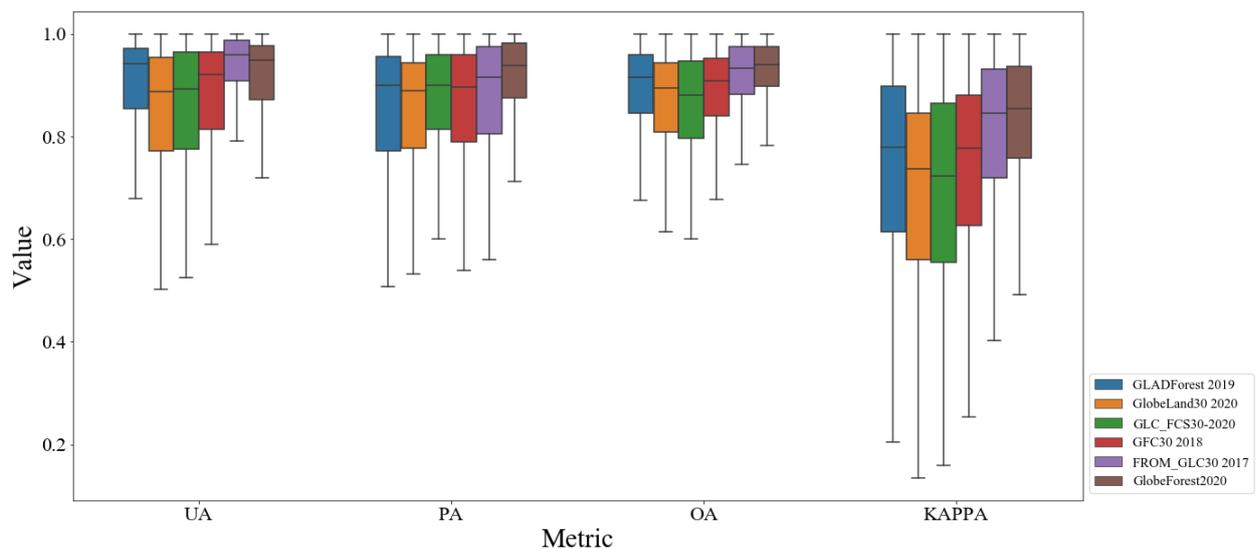

Figure 21 Box plots for each product.

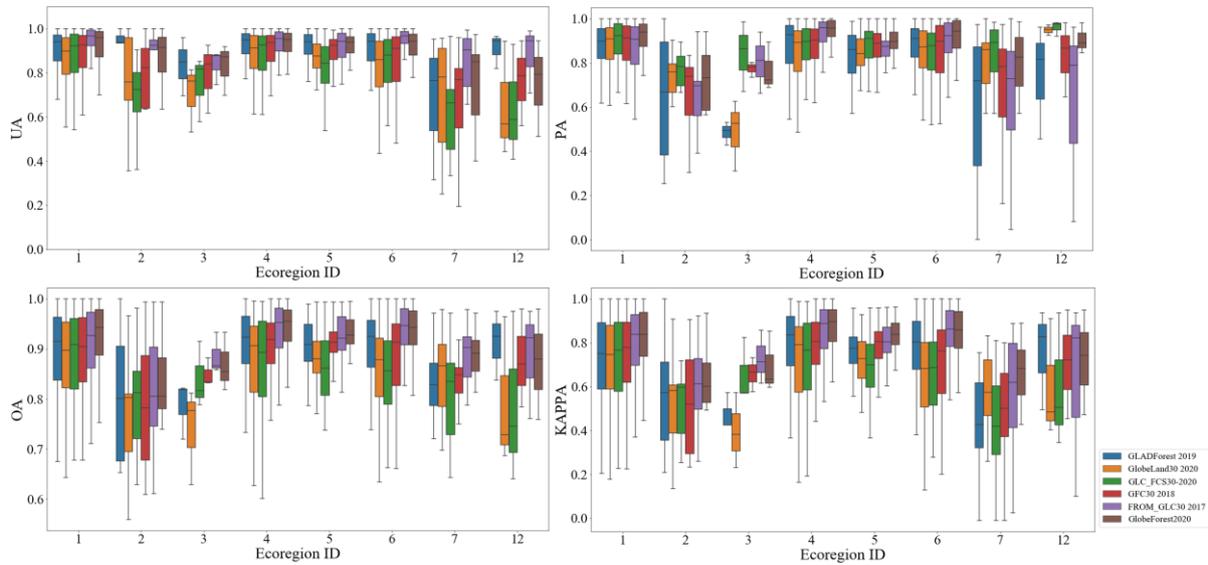

Figure 22 Box plots for each product in different ecoregions ( Note: Ecoregion ID and corresponding ecoregion: ID 1 is Tropical and Subtropical Moist Broadleaf Forests, ID 2 is Tropical and Subtropical Dry Broadleaf Forests, ID 3 is Tropical and Subtropical Coniferous Forests, ID 4 is Temperate Broadleaf and Mixed Forests, ID 5 is Temperate Conifer Forests, ID 6 is Boreal Forests/Taiga, ID 7 is Tropical and Subtropical Grasslands, Savannas and Shrublands, ID 12 is Mediterranean Forests, Woodlands, and Scrub)

In uncertain grids, certain samples and uncertain samples are adopted to achieve the best accuracy which has been proved in section 4.1. It is worth noting that in certain regions we only use certain samples as input to the random forest classifier without uncertain samples, considering that pixels are more certain to be classified in these regions, and still obtain the highest precision. GlobeForest2020 has been proved to exceed the previous highest state-of-the-art overall accuracies (obtained by Gong et al., 2017) by 2.77% in uncertain grids and by 1.11 % in certain grids. Accuracies of 6 products in certain and uncertain regions are shown in Table 10.

Table 10 Accuracies of global forest products in certain and uncertain grids.

| Product | PAcer | UAcer | OAcer | KAPPAcer | PAunc | UAunc | OAunc | KAPPAunc |
|---|---|---|---|---|---|---|---|---|
| FROM_GLC30 2017 | 0.8584 | **0.9224** | 0.8941 | 0.7880 | 0.7508 | **0.9361** | 0.8469 | 0.6948 |
| GFC30 2018 | 0.8117 | 0.8518 | 0.8367 | 0.6733 | 0.7902 | 0.8215 | 0.8056 | 0.6113 |
| GLADForest 2019 | 0.8164 | 0.8446 | 0.8346 | 0.6691 | 0.7832 | 0.8356 | 0.8110 | 0.6223 |
| GLC_FCS30-2020 | 0.8463 | 0.8108 | 0.8260 | 0.6521 | 0.8414 | 0.7405 | 0.7689 | 0.5363 |
| GlobeLand30 2020 | 0.7405 | 0.8170 | 0.7892 | 0.5781 | 0.7204 | 0.8041 | 0.7680 | 0.5368 |
| GlobeForest2020 | **0.9157** | 0.8953 | **0.9052** | **0.8104** | **0.8626** | 0.8881 | **0.8746** | **0.7492** |

Note: PAcer: PA of certain grids, UAcer: UA of certain grids, OAcer: OA of certain grids, KAPPAcer: KAPPA of certain grids, PAunc: PA of uncertain grids, UAunc: UA of uncertain grids, OAunc: OA of uncertain grids, KAPPAunc: KAPPA of uncertain grids.

Table 11 Global accuracy of six products.

| Product | PA | UA | OA | KAPPA |
|---|---|---|---|---|
| FROM_GLC30 2017 | 0.8029 | **0.9290** | 0.8700 | 0.7403 |
| GFC30 2018 | 0.8006 | 0.8361 | 0.8209 | 0.6418 |
| GLADForest 2019 | 0.7993 | 0.8400 | 0.8226 | 0.6453 |
| GLC_FCS30-2020 | 0.8438 | 0.7730 | 0.7969 | 0.5936 |
| GlobeLand30 2020 | 0.7301 | 0.8103 | 0.7784 | 0.5571 |
| GlobeForest2020 | **0.8883** | 0.8917 | **0.8896** | **0.7792** |

To intuitively compare the accuracy of GlobeForest2020 and other products, let us consider the scatter plots representing the accuracy of one previous product (corresponding to the x axes) and GlobeForest2020 (corresponding to the y axes) in unit of 5° grid (Figure 23). Most of the points (i.e., 5°grid) are distributed above the line with slope of 1, whose values of y-coordinate are larger than values of x-coordinate, indicating that GlobeForest2020 has higher accuracy than the other products in most 5° grids. The fitting line in Figure 23 (b) has the smallest intercept, the largest slope, and the highest $R^2$, indicating that GlobeForest2020 and FROM_GLC30 2017 have the closest accuracy. The fitting line in Figure 23 (a) has the largest intercept and the smallest slope, indicating that GlobeForest2020 and globeLand30 2020 have the largest accuracy differences. Therefore, among the 5 forest products, FROM_GLC30 2017 has the highest accuracy, and GlobeLand30 2020 has the lowest accuracy, which is consistent with the conclusion in Table 11.

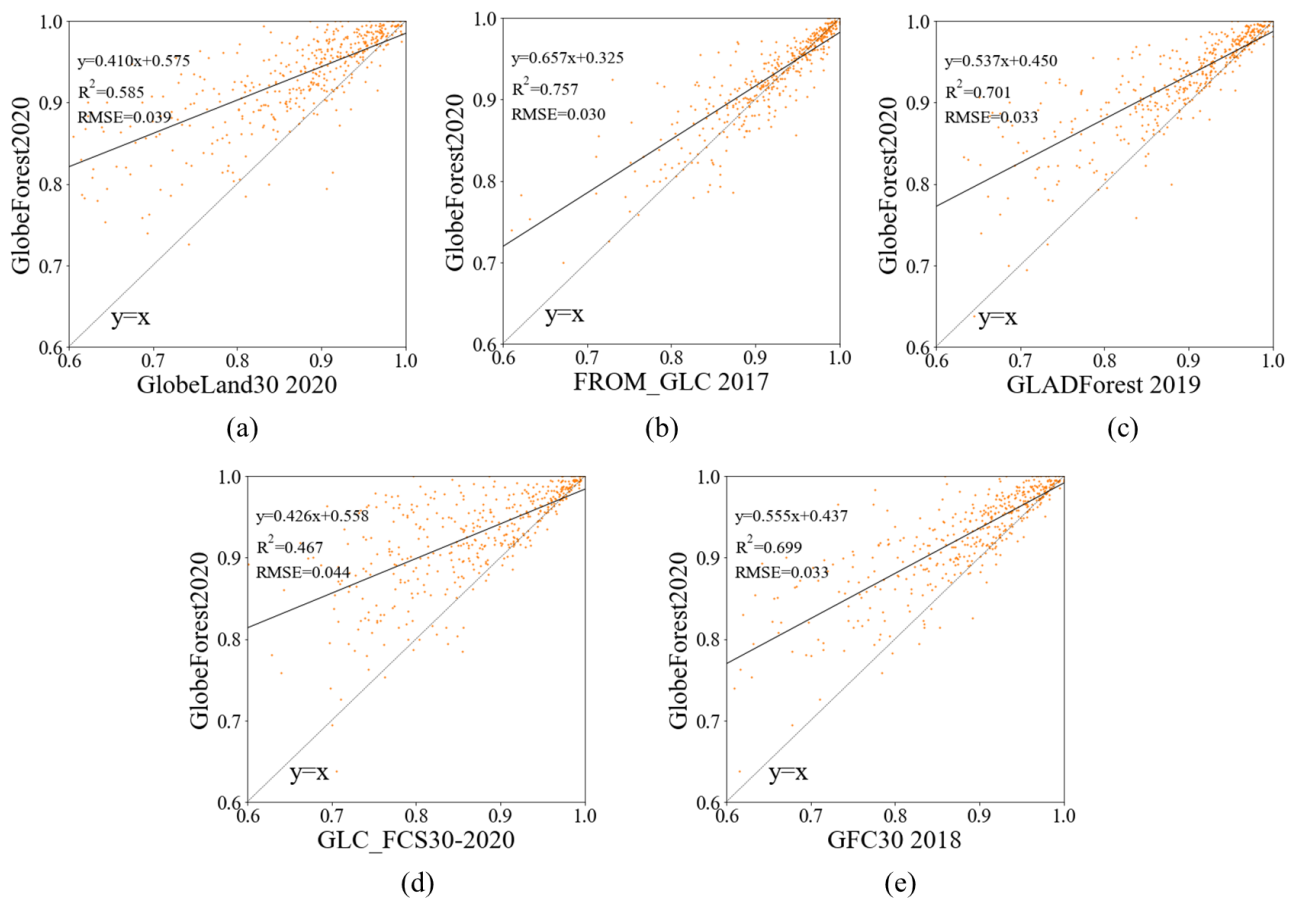

Figure 23 Accuracy relationship between GlobeForest2020 and other products.

Let us now focus our attention on the geographical distribution of the accuracy of the considered GFC products. In Figure 24, the accuracy degree is mapped from high to low by means of a color map that spans from green to red, respectively. FROM_GLC30 2017 and GlobeForest2020 show high spatial resemblance and most green grids are distributed in these two maps. Nonetheless, they vary in Africa, since FROM_GLC 2017 outperforms GlobeForest2020 except for North America and North Asia. The biggest differences of 6 products' accuracy are found in Europe, Asia, and North America, where more red and yellow grids are shown in GlobeLand30 2020 and GLC_FCS30-2020, while more green grids are shown in FROM_GLC 2017 and GlobeForest2020. Comparing with GLC_FCS30-2020 and GlobeLand30 2020, GLADForest 2019 and GFC30 2018 have higher spatial variations in accuracies. Distribution of grids with accuracy higher than 0.9 is similar with that of GlobeForest2020. However, they have more grids with accuracy below 0.8 than that of GlobeForest2020. Among all the continents, spatial uniformity of products' accuracies occurs in South America, where the high-accuracy grids mainly distribute in the central zone where the Amazon Forest is located and southern regions.

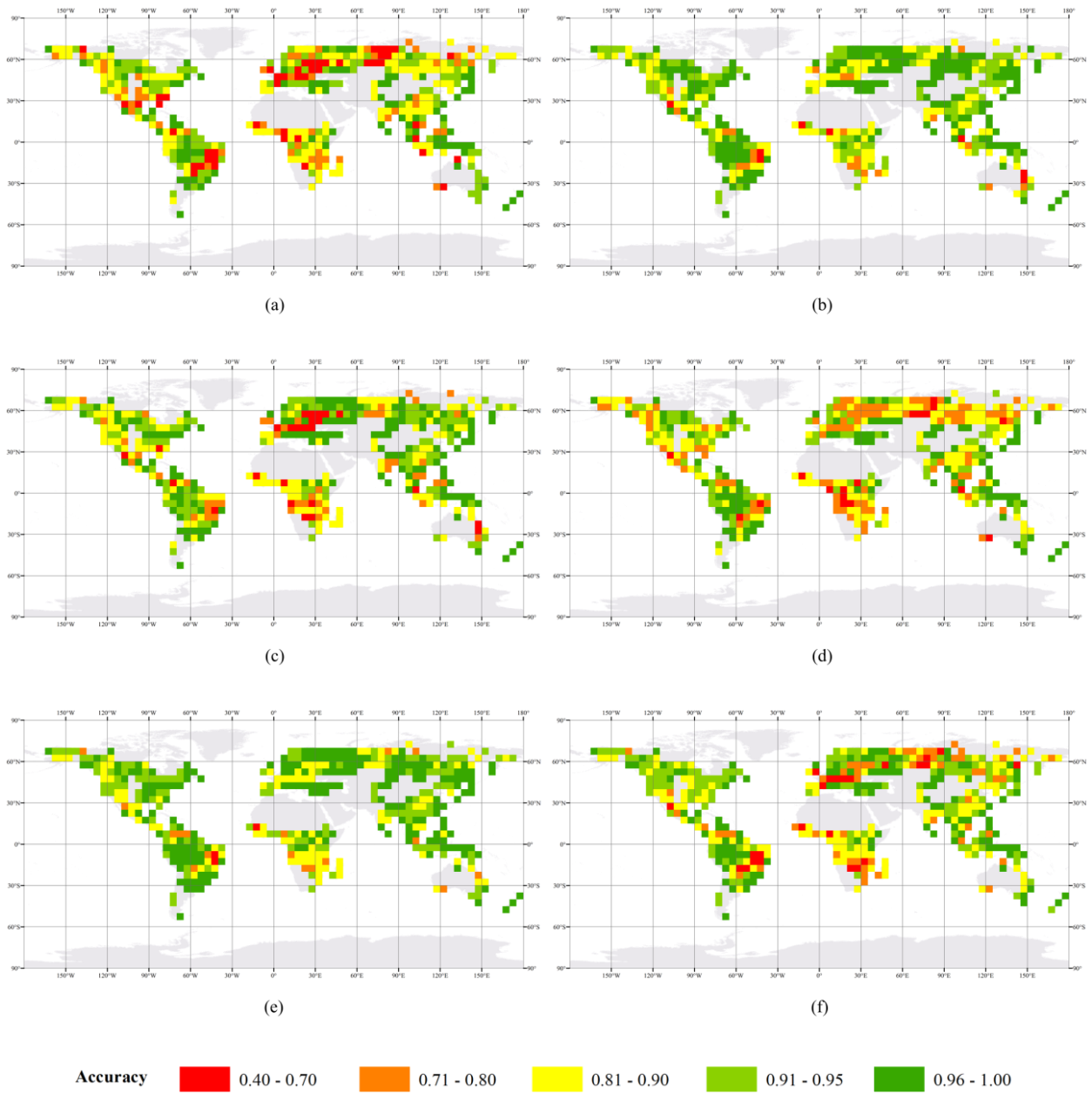

Figure 24 Overall accuracy of 6 products in unit of 5°grid. (a) GlobeLand30 2020. (b) FROM_GLC30 2017. (c) GLADForest 2019. (d) FCS_GLC30-2020. (e) GlobeForest2020 (ours). (f) GFC30 2020.

## V. Discussion

In this section, the effectiveness of semi-automatic annotating mechanism, classification sampling strategies, and advantages of GlobeForest2020 are discussed as follows.

## 5.1 Semi-automatic Annotating Mechanism

The main design idea of semi-automatic annotating mechanism is to reduce the manual labor by finding more consistent samples by multi-classifiers' voting results. Meanwhile, at least 25 predict labels from different classifiers trained with different training sample sets, different dimensional features and different classification models are participated into voting, the predict overall accuracy of consistent samples could be up to 98% when the consistent ratio reaches 0.9. And the newly manually labeled samples can be added to retrain the classifier in a new iteration. Particularly, the samples with wrong label revised by the human could greatly improve the accuracy of each meta classifier. As a result, more samples could be regarded as consistent samples, and the labeling costs could be further reduced.

The whole labeling mechanism yielded 23% inconsistent samples and 77% consistent samples from the approximately 400,000 samples needing annotation in this paper. To keep high quality of the sample set, we assumed each inconsistent sample should be labeled 3 times, and consistent samples should be labeled once, hence 51.3% (616,000 times in 1,200,000 times) manual annotation could be reduced. If we could relax the extreme quality requirement, the consistent samples could be directly used as training samples, 77% (924,000 times in 1,200,000 times) manual annotations could be reduced.

## 5.2 The Future Application of FSS

FSS is a manually annotated data set with geo-coordinate and corresponding HR-GEIs for labeling, which contains 395280 scattered samples categorized as forest, shrubland, grassland, impervious surface in the global scale. In this paper, we only used classic RF classifier trained with spectral features by utilizing part samples in FSS. In future work, more elaborate classifier from the spectral-spatial-temporal feature fusion and ensemble model could be designed to further improve the classification accuracy based on the FSS.

Furthermore, FSS could provide an effective training sample set to generate long time series land cover products, which is still an unsolved problem due to the domain discrepancy between images obtained from different years. FFS can be expanded in other images acquired in the previous years by change analysis between images obtained by previous year and 2020. Geo-points in FFS with unchanged labels can be new samples in the previous images used to train new classifier in that year. In result, a separate classifier would be trained for each year, and the problem of domain discrepancy between different years can be bypassed.

## 5.3 Generation of GlobeForest2020

The global forest classification map was improved from the view of sampling strategy in this paper. We firstly evaluated the accuracy of certain samples which voted by 0, 4 and 5 times by five GFC products, and the median values of their confidence levels were 92%, 92% and 97%, respectively. The utilization of free training samples can keep the robustness of classifier in each 5° grid for adequate training sample source. In certain 5° grids, only free training samples could get satisfied accuracy. In uncertain 5° grids, combing uncertain labeled samples can substantially improve the classification accuracy. Based on FSS, comparison of five sampling strategies revealed that it is not the case that more samples are more effective. On the contrary, sampling strategy is more important for better classification performance.

As forest cover change mainly occurs in transition zones with high classification uncertainty, GlobeForest2020 promotes accuracy in uncertain regions which is essential not only for more accurate forest mapping but also for forest change and disturbance monitoring. Therefore, based on GlobeForest2020, we will continue to work for generation of long time-series forest cover change products, which is vital for analyzing the change trend and disturbance of global forest.

## 5.4 Comparison of GlobeForest2020 and Other Products

Comparing GlobeForest2020 and other GFC products, GlobeForest2020 performed best for each continent, followed by FROM_GLC30 2017 whose accuracies of 5° grids had the highest relevance with those of GlobeForest2020. Comparing the two sets of products, GlobeForest2020 performed better than FROM_GLC30 2017, except in parts of Africa where the accuracy was slightly lower than FROM_GLC30 2017. From the view of area statistics, GlobeForest2020 had the highest forest area, in line with the conclusion that although the UA was 1.14% lower than the highest value, the PA outperformed the second highest value by 2.31%. As for a comparison of forest layers of three sets of GLC products, namely FROM_GLC 2015, GLC_FCS30-2015 and GlobeLand30-2010, the accuracy validation of the latest yearly distribution of the three products in this paper found that the forest layer accuracy of FROM_GLC30 2017 was significantly better than those of GLC_FCS30-2020 and GlobeLand30 2020. As for the classification methods, GLC_FCS-2020, FROM_GLC30 2017 were generated using supervised classification models, while GlobeLand30 2020 was produced with an object-based segmentation method which probably caused larger omission error and was detrimental to PA. As for the two sets of GFC products, GLADForest 2019 and GFC30 2018 had similar accuracies and performed better than GLC_FCS30-2020 and GlobeLand30 2020, but did not outperform FROM_GLC30 2017.

# VI. Conclusion and Future work

In this study, review and comparison of previous global forest products show that there are still many regions with relatively low agreement, which are found to play a critical role in ecosystem. Therefore, a third-party evaluation of these products, is necessary to improve the application value of GFC products. However, annotating forest sample sets is time consuming and laborious, which significantly reduces the speed of validation sample set generation. To solve this problem, we build a semi-automatic sample labeling mechanism that actively annotates most inconsistent sample set with low manpower cost and high labeling quality. Based on the labeling mechanism above, 395280 scattered samples distributed around the global were labeled, mainly including shrub, grassland, and cropland. Using these labeled samples, we fairly evaluated exiting forest cover products and reported accuracy for each 5°×5° geographical grid cell and produced a more accurate classification map (GlobeForest2020) using the optimal sampling strategy.

Comparative experiments were designed to investigate which combination of samples achieves the best classification accuracy when used as input to the classifier. For uncertain regions, combing uncertain sample points can substantially improve the classification accuracy. However, it is not the case that more samples are more effective. On the contrary, sampling strategy is more important for better classification performance. A classifier constructed with a smaller number of local samples from the grid to be classified and the ecoregions it locates in is better than a large number samples from global scale. Result showed that our product achieves the best accuracy among all 5 forest products. Therefore, it is concluded that GlobeForest2020 is a promising accurate forest product which can enrich the existing forest products and provide significant support for forest cover monitoring.

Our future work will aim at further utilization of existing HR-GEI images and land-cover with all class products generation. Firstly, HR-GEI patches having more than one classes of land covers are not identified and annotated, resulting in wasteful use of images. For full use of images, we will introduce an attention mechanism to precisely correspond the category to a point in the HR-GEI patches. Secondly, the forest classification method we adopted in this paper has been proven to be effective, and we will apply the method to generate global land cover products.

## VII. Data Availability

The forest product of GlobeForest2020 generated in this paper is available at Google Earth Engine (GEE) code platform with URL: https://code.earthengine.google.com/42cc67561056b1f0b331e372917ea43b. The product is grouped by 1137 5°×5° regional tiles with resolution of 30 m. Each image contains a label band with values of 0 or 1. Value of 0 represents non-forest and 1 represents forest. The manually labeled samples are published on GEE platform with URL: https://code.earthengine.google.com/?asset=users/410093033/Lat_Lon_Type_Filename. The HR-GEI with filename including geohash which encodes a geographic location can be acquired with the Google Drive URL: https://drive.google.com/drive/folders/1PS6PdvxHCliSCOpYwt0Pjwkk_VuH6zTM?usp=sharing.

## Acknowledgements

This work was supported in part by the National Key R&D Program of China under Grant 2019YFA0607203; in part by the National Natural Science Foundation of China under Grant 61976234; in part by the Guangdong Basic and Applied Basic Research Foundation under Grant 2019A1515011057; in part by the Guangzhou Basic and Applied Basic Research Project; in part funded by Centre for Integrated Remote Sensing and Forecasting for Arctic Operations (CIRFA) and the Research Council of Norway (RCN Grant no. 237906).

Dorren, L. K., Maier, B., & Seijmonsbergen, A. C. (2003). Improved Landsat-based forest mapping in steep mountainous terrain using object-based classification. *Forest Ecology and Management, 183*(1-3), 31-46.

Friedl, M. A., Sulla-Menashe, D., Tan, B., Schneider, A., Ramankutty, N., Sibley, A., & Huang, X. (2010). MODIS Collection 5 global land cover: Algorithm refinements and characterization of new datasets. *Remote sensing of Environment, 114*(1), 168-182.

Gong, P., Liu, H., Zhang, M., Li, C., Wang, J., Huang, H., . . . Song, L. (2019). Stable classification with limited sample: transferring a 30-m resolution sample set collected in 2015 to mapping 10-m resolution global land cover in 2017. *Science Bulletin, 64*(6), 370-373. doi:https://doi.org/10.1016/j.scib.2019.03.002

Gong, P., Wang, J., Yu, L., Zhao, Y., Zhao, Y., Liang, L., . . . Liu, S. (2013). Finer resolution observation and monitoring of global land cover: First mapping results with Landsat TM and ETM+ data. *International Journal of Remote Sensing, 34*(7), 2607-2654.

Hagner, O., & Reese, H. (2007). A method for calibrated maximum likelihood classification of forest types. *Remote sensing of environment, 110*(4), 438-444.

Hansen, M. C., DeFries, R. S., Townshend, J. R., & Sohlberg, R. (2000). Global land cover classification at 1 km spatial resolution using a classification tree approach. *International journal of remote sensing, 21*(6-7), 1331-1364.

Hansen, M. C., Potapov, P. V., Moore, R., Hancher, M., Turubanova, S. A., Tyukavina, A., . . . Loveland, T. R. (2013). High-resolution global maps of 21st-century forest cover change. *science, 342*(6160), 850-853.

He, K., Zhang, X., Ren, S., & Jian, S. (2016). *Deep Residual Learning for Image Recognition.* Paper presented at the IEEE Conference on Computer Vision & Pattern Recognition.

Hu, J., Shen, L., & Sun, G. (2018). *Squeeze-and-excitation networks.* Paper presented at the Proceedings of the IEEE conference on computer vision and pattern recognition.

Jia, T., Li, Y., Shi, W., & Zhu, L. (2019). Deriving a forest cover map in Kyrgyzstan using a hybrid fusion strategy. *Remote Sensing, 11*(19), 2325.

Jiyuan, L. J. o. R. S. (1997). Study on National Resources & Environment Survey and Dynamic Monitoring Using Remote Sensing [J]. *3*.

Jun, C., Ban, Y., & Li, S. (2014). Open access to Earth land-cover map. *Nature, 514*(7523), 434-434. doi:10.1038/514434c

Kasetkasem, T., Arora, M. K., & Varshney, P. K. (2005). Super-resolution land cover mapping using a Markov random field based approach. *Remote Sensing of Environment, 96*(3-4), 302-314.

Knauer, U., von Rekowski, C. S., Stecklina, M., Krokotsch, T., Pham Minh, T., Hauffe, V., . . . Chmara, S. (2019). Tree species classification based on hybrid ensembles of a convolutional neural network (CNN) and random forest classifiers. *Remote Sensing, 11*(23), 2788.

Lamarche, C., Santoro, M., Bontemps, S., d'Andrimont, R., Radoux, J., Giustarini, L., . . . Arino, O. (2017). Compilation and validation of SAR and optical data products for a complete and global map of inland/ocean water tailored to the climate modeling community. *Remote Sensing, 9*(1), 36.

Li, W., Dong, R., Fu, H., Wang, J., Yu, L., & Gong, P. (2020). Integrating Google Earth imagery with Landsat data to improve 30-m resolution land cover mapping. *Remote Sensing of Environment, 237*, 111563.

Li, X., Du, Y., & Ling, F. (2013). Super-resolution mapping of forests with bitemporal different spatial
56